\documentclass[10pt,twocolumn,letterpaper]{article}

\usepackage[]{main}   
\definecolor{iccvblue}{rgb}{0.21,0.49,0.74}
\usepackage[pagebackref,breaklinks,colorlinks,allcolors=iccvblue]{hyperref}

\usepackage{amsmath,amssymb,amsfonts}
\usepackage{dsfont}
\usepackage{multirow}

\usepackage{booktabs} 
\usepackage{setspace} 
\usepackage{float} 
\usepackage{makecell}
\usepackage{threeparttable}
\usepackage{tablefootnote}

\title{
S5: Scalable Semi-Supervised Semantic Segmentation in Remote Sensing
}

\author{
Liang Lv$^1$,
Di Wang$^{1,2}$,
Jing Zhang$^1$\thanks{Corresponding author.},
Lefei Zhang$^1$\footnotemark[1] \\
$^1$National Engineering Research Center for Multimedia Software, \\
School of Computer Science, Wuhan University \\
$^2$Zhongguancun Academy \\
{\tt\small lianglyu@whu.edu.cn, d\_wang@whu.edu.cn, jingzhang.cv@whu.edu.cn, zhanglefei@whu.edu.cn}
}

\begin{document}
\maketitle

\begin{figure*}[t]
\centering
\includegraphics[width=\linewidth]{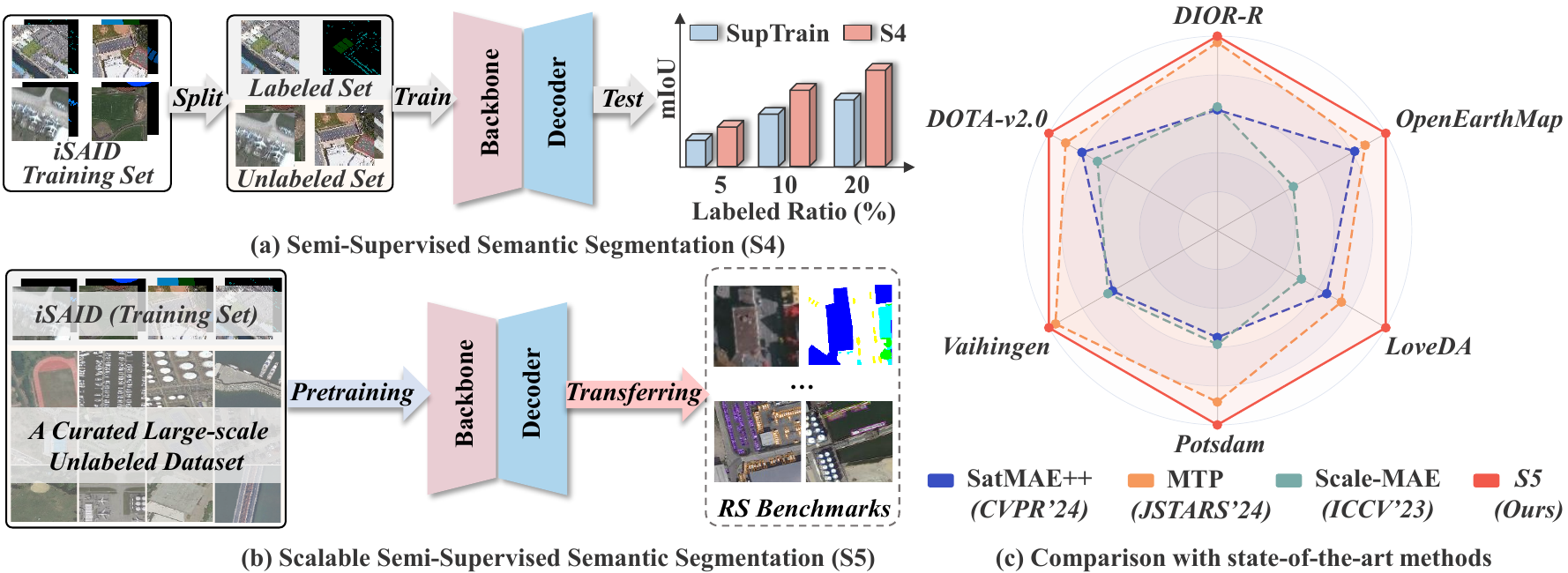}
\caption{(a) Traditional S4 workflow: splitting the dataset into labeled and unlabeled subsets to improve model performance with few labeled samples.  
(b) The proposed S5 workflow: perform semi-supervised segmentation pretraining on both labeled and large-scale unlabeled datasets, followed by fine-tuning on RS benchmarks. 
(c) Comparison of performance across four RS segmentation and two object detection benchmarks.}
\label{fig:method_overview}
\end{figure*}

\begin{abstract}
Semi-supervised semantic segmentation (S4) has advanced remote sensing (RS) analysis by leveraging unlabeled data through pseudo-labeling and consistency learning. However, existing S4 studies often rely on small-scale datasets and models, limiting their practical applicability.  To address this, we propose S5, the first scalable framework for semi-supervised semantic segmentation in RS, which unlocks the potential of vast unlabeled Earth observation data typically underutilized due to costly pixel-level annotations. Built upon existing large-scale RS datasets, S5 introduces a data selection strategy that integrates entropy-based  filtering and diversity expansion, resulting in the RS4P-1M dataset. Using this dataset, we systematically scale up S4  into a new pretraining paradigm, S4 pre-training (S4P), to pretrain RS foundation models (RSFMs) of varying sizes on this extensive corpus, significantly boosting their performance on land cover segmentation and object detection tasks. Furthermore, during fine-tuning, we incorporate a Mixture-of-Experts (MoE)-based multi-dataset fine-tuning approach, which enables efficient adaptation to multiple RS benchmarks with fewer parameters. This approach improves the generalization and versatility of RSFMs across diverse RS benchmarks. The resulting RSFMs achieve state-of-the-art performance across all benchmarks, underscoring the viability of scaling semi-supervised learning for RS applications. All datasets, code, and models will be released at \href{https://github.com/MiliLab/S5}{S5}.
\end{abstract}

\section{Introduction}
\label{sec:intro}

Remote sensing (RS) semantic segmentation is a key task in RS image understanding, aiming to accurately classify each pixel to enable automatic recognition and analysis of land cover information \cite{zhang2022artificial}. However, traditional fully supervised segmentation methods heavily rely on pixel-level annotations, making the acquisition of high-quality training samples extremely costly. This dependency also limits model generalization in diverse scenarios. To reduce the burden of manual labeling and lower costs, semi-supervised semantic segmentation (S4) \cite{CCT} has gained increasing attention. S4 enhances RS image segmentation performance by combining a small amount of labeled images with a large number of unlabeled images during the training stage.

Early S4 research explored GAN-based methods \cite{semiGAN}. Subsequently, data augmentation was recognized as a crucial factor \cite{need}. Recent approaches mainly rely on pseudo-labeling and consistency regularization. ST++ \cite{ST++} demonstrates that strong data augmentation significantly boosts self-training performance, though multi-stage pipelines often reduce efficiency. UniMatch \cite{unimatch} revisits consistency regularization using weak-to-strong augmentation, a strategy originally generalized by FixMatch \cite{fixmatch} for semi-supervised classification. FixMatch generates pseudo-labels on weakly augmented images and enforces consistency on their strongly augmented counterparts, using a confidence threshold to ensure pseudo-label reliability. Thanks to its simplicity and efficiency, FixMatch has become a key baseline in S4, inspiring many follow-up methods, such as UniMatch \cite{unimatch}, RankMatch \cite{RankMatch}, and CorrMatch \cite{CorrMatch}. In RS, methods like RanPaste \cite{RanPaste}, WSCL \cite{WSCL}, and SegMind \cite{SegMind} also adopt the FixMatch framework, enhancing it with domain-specific augmentations such as random copy-paste, dual-view augmentation, and random masking for complex RS scenes.

 However, current S4 research still relies on small-scale models and datasets. As illustrated in Figure \ref{fig:method_overview} (a), a common strategy involves splitting the training set of standard segmentation datasets (e.g., iSAID \cite{iSAID}) into labeled and unlabeled subsets. By leveraging S4 methods in combination with unlabeled images, model performance under limited supervision can be significantly improved compared to purely supervised training (SupTrain). However, such settings are typically constrained to a single dataset, which limits the exploration of S4's potential in harnessing large-scale Earth observation data.

Meanwhile, RS foundational models (RSFMs) have made significant progress, benefiting from large datasets like MillionAID \cite{MillionAID} and SAMRS \cite{SAMRS}, alongside extensive exploration of different pre-training strategies. Self-supervised learning methods, such as contrastive learning \cite{contrastive} and masked image modeling (MIM) \cite{MIM}, extract generalizable features without relying on labeled data. In contrast, supervised pre-training better aligns upstream and downstream tasks and domains, enhancing transferability. For instance, RSP \cite{RSP} applies supervised pre-training on the MillionAID dataset, producing RSFMs that perform well across diverse downstream tasks. The SAMRS \cite{SAMRS} dataset is used to explore segmentation pre-training (SEP) and multitask pre-training (MTP) \cite{MTP}, which further enhance generalization by narrowing the gap between pre-training and target tasks. However, SAMRS relies on the Segment Anything Model (SAM) \cite{SAM} to generate segmentation masks from bounding-box annotations. This dependence, along with its limited annotation scale constrains scalability. In particular, this mask generation process resembles the self-training paradigm of S4. Furthermore, given that pre-training is most effective when tasks and domains are well aligned, a key question arises: \textit{Can we scale up S4 to pre-train RSFMs on massive RS imagery, and thereby enhance their performance across diverse RS applications?}

To answer this question, we introduce Scalable Semi-supervised Semantic Segmentation (S5), the first framework to pre-train RSFMs using large-scale unlabeled RS images via semi-supervised learning. 

As illustrated in Figure \ref{fig:method_overview} (b), S5 curates a large-scale unlabeled RS dataset using a sample selection strategy that combines entropy-based filtering with diversity expansion. Building on this dataset, we systematically perform S4 pre-training (S4P) on RSFMs of varying capacities, 
all initialized with MAE \cite{MIM} pretrained weights. 
Experiments show that S4P not only enhances the general representation ability learned from MAE pre-training, but also significantly improves downstream performance on land cover segmentation and object detection tasks. Furthermore, during fine-tuning, S5 introduces a multi-dataset fine-tuning strategy based on Mixture-of-Experts (MoE), which incorporates both task-shared and task-specific Feedforward Networks (FFNs). This design allows the RSFM to adapt efficiently to multiple RS benchmarks with minimal additional parameters. Extensive experiments demonstrate that RSFMs pre-trained under the S5 framework achieve state-of-the-art (SOTA) performance across multiple segmentation and detection benchmarks (see Figure \ref{fig:method_overview} (c)), validating the potential and promise of large-scale semi-supervised pre-training for RSFMs development.

Our main contributions are summarized as follows:

\begin{itemize}
 \item We introduce the S5 framework for RS, addressing the limitations of traditional S4 methods that rely on small-scale datasets and models. S5 establishes a new paradigm for leveraging vast amounts of unlabeled RS images to develop RSFMs.

\item We propose a low-entropy filtering and diversity expansion strategy to curate RS4P-1M, a dataset of one million reliable RS images covering diverse geospatial scenes. This dataset enables effective S4P of RSFMs and lays a solid foundation for future research in scalable RSFMs pre-training.

\item We design a MoE-based multiple dataset fine-tuning (MoE-MDF) approach, which combines task-shared and task-specific FFNs to enable efficient joint adaptation across multiple RS benchmarks with minimal parameter overhead, significantly enhancing RSFMs' generalization and transferability.
 \end{itemize}
\section{Related Work}
\label{sec:formatting}

\begin{figure*}[t] 
\centering
\includegraphics[width=\textwidth]{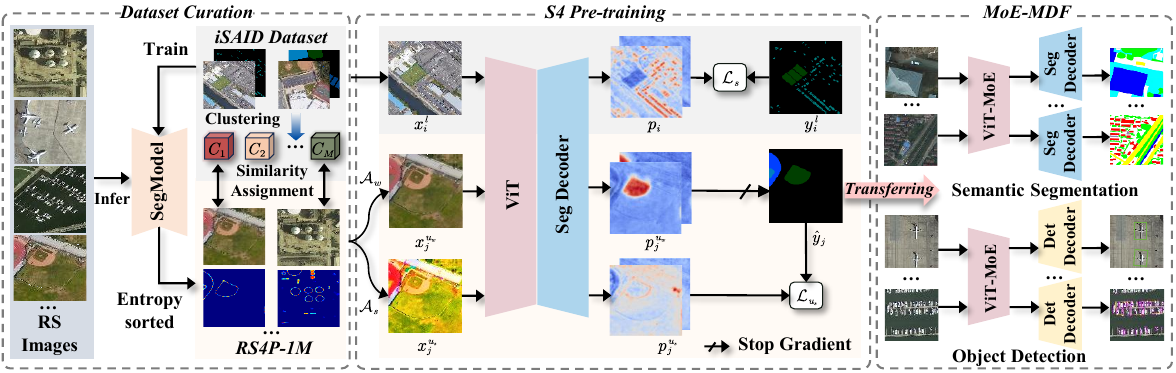}
\caption{The overall pipeline of the proposed S5 framework. It starts with the construction of the RS4P-1M dataset, followed by training RSFMs  based on the S4P. The pre-trained weights are then fine-tuned on semantic segmentation and object detection benchmarks through the MoE-based multiple dataset fine-tuning (MoE-MDF) scheme. ViT-MoE indicates the integration of the FFN-MoE modules into the ViT backbones.}
\label{fig:overview}
\end{figure*}

\subsection{Semi-supervised Semantic Segmentation}

S4 aims to train semantic segmentation models using a small amount of labeled data and a large pool of unlabeled data. Early S4 approaches relied on consistency regularization and pseudo-labeling, ensuring stable predictions under perturbations or leveraging the model's own outputs as labels. Recent deep learning-based S4 methods have significantly improved performance. UniMatch \cite{unimatch} enforces weak-to-strong consistency at both image and feature levels, AugSeg \cite{AugSeg} enhances robustness through data augmentation, and iMAS \cite{iMAS} introduces instance-specific, model-adaptive supervision. CorrMatch \cite{CorrMatch} propagates labels via correlation matching, AllSpark \cite{wang2024allspark} employs a Transformer-based approach leveraging labeled features, SemiVL \cite{SemiVL} integrates vision-language guidance for better pseudo-labeling, and UniMatchV2 \cite{UnimatchV2}, built on DINOV2 \cite{DINOV2}, further improves S4 performance. 

S4 methods in RS address domain-specific challenges. For instance, RanPaste \cite{RanPaste} exploits unlabeled data through consistency and pseudo-labeling, WSCL \cite{WSCL} transitions from weak to strong labels, SegMind \cite{SegMind} integrates mask image modeling with contrastive learning, and DWL \cite{DWL} enhances segmentation via decoupling and weighting of different components. Unlike existing S4 methods, which are often constrained by small datasets, we introduce S5, a scalable semi-supervised framework designed to leverage large-scale unlabeled RS data for pre-training RSFMs.

\subsection{Remote Sensing Foundation Models} 

RSFMs have gained attention for their ability to learn generalizable and transferable features. Pre-training approaches are typically supervised or self-supervised. RSP \cite{RSP} pioneered this approach by pre-training CNNs and vision transformers on Million-AID \cite{MillionAID}, while MTP \cite{MTP} aligns multiple downstream tasks for enhanced representation learning. SAMRS \cite{SAMRS} improves segmentation-focused FMs using a 100,000-sample dataset built with SAM \cite{SAM}. However, these methods rely on labeled data, which are scarce and hard to scale. Recent work favors self-supervised pre-training, either contrastive—forming positive-negative pairs from image augmentations \cite{Global}, multimodal images \cite{multiview,multimodal}, geographic priors \cite{Geographical}, or temporal imagery \cite{temporal}—or generative methods like MIM \cite{MIM} that reconstruct masked regions to capture structural features. RingMo \cite{RingMo} employs incomplete masking for dense small objects in RS scenes, while RVSA \cite{RVSA}, initialized with MIM weights, introduces rotational window attention to enhance target representation with lower computational cost. GFM \cite{GFM} refines MIM pre-training by leveraging ImageNet-pre-trained FMs, while SatMAE \cite{SatMAE} and Scale-MAE \cite{Scale-MAE} incorporate multi-temporal and multi-scale features, respectively. SkySense \cite{SkySense} proposes a billion-scale multi-modal RSFM that leverages multi-level contrastive learning and geo-context learning for large-scale pre-training on RS images. In contrast, our work explores S4P for RSFMs and introduces S5: the first scalable framework for semi-supervised semantic segmentation in RS. Leveraging the newly established  RS4P-1M dataset in this work, we successfully pre-train RSFMs with up to 600M parameters, achieving SOTA performance across multiple benchmarks.
\section{Methodology}

As illustrated in Figure~\ref{fig:overview}, the proposed S5 framework consists of three key components: dataset curation, S4 pre-training (S4P), and MoE-based multi-dataset fine-tuning (MoE-MDF). These components will be introduced in detail in the following text.

\subsection{Pre-training Dataset Curation}

To explore a suitable S4P paradigm for large-scale RS images, we begin by analyzing existing public datasets. In RS, the MillionAID dataset \cite{MillionAID} is one of the most widely used resources for pre-training RSFMs, due to its large scale and diverse scene types. As such, it serves as our primary source of unlabeled imagery. To enhance dataset diversity, we further incorporate SAMRS \cite{SAMRS} and STAR \cite{STAR} datasets. As these datasets are primarily collected from Google Earth, we adopt iSAID \cite{iSAID}, an object-level segmentation dataset with similar resolution and imaging style, as our labeled data source.

Based on the above datasets, we combine a certain amount of labeled data with large-scale unlabeled images, and leverage S4P to transfer annotation knowledge to broader RS scenarios. This enhances the model's generalization ability and bridges the gap between pre-training and downstream tasks. However, using all unlabeled data may degrade pseudo-label quality due to distribution mismatch. 

To mitigate this, we propose an entropy-based filtering and semantic diversity expansion strategy. Treating the labeled set as a curated subset, we aim to extend it with unlabeled samples that are both reliable and diverse. We first train an initial segmentation model (ViT-H + UperNet \cite{UperNet}) on the labeled data and apply it to cropped unlabeled patches. Considering that the model may exhibit high uncertainty on low-quality or out-of-distribution samples, we adopt the pixel-level average entropy as a confidence measure for the pseudo-labels. For each unlabeled image $x \in \mathbb{R}^{H \times W \times 3}$, the average entropy is defined as:
\begin{equation}
E(x) = -\frac{1}{H \times W} \sum_{i=1}^{H \times W} \sum_{k=1}^{K} P^k(x^i) \log P^k(x^i),
\end{equation}
where $P^k(x^i)$ is the probability of pixel $x^i$ belonging to class $k$ (excluding background). Patches are then ranked by entropy:
\begin{equation}
\begin{aligned}
\mathcal{D} = \left\{ \left( x_j, E_j \right),\ j = 1, \dots, N \right\}, \\
\text{with } E_1 \leq E_2 \leq \cdots \leq E_N.
\end{aligned}
\end{equation}
where $N$ is the total number of cropped unlabeled image patches. High-entropy samples are likely noisy or out-of-distribution, so we prioritize low-entropy patches. Nevertheless, low entropy selection may yield semantic redundancy, with the chosen images concentrated in a few common scenes. To promote diversity, 
as illustrated in the similarity assignment step of Figure \ref{fig:overview},
we extract features from labeled images using the trained segmentation model’s backbone and apply K-Means to cluster them into M clusters, denoted as $C_1, C_2, \ldots, C_M$. For each unlabeled image, we compute its cosine similarity to these cluster prototypes and assign it to the nearest cluster. Let $B^u$ denote the target number of selected unlabeled samples, where $N > B^u$. Each cluster is allocated a quota:
\begin{equation}
B_m^u = B^u \cdot \frac{N_m^l}{B^l},
\end{equation}
$N_m^l$ denotes the number of labeled samples in the $m$-th cluster, and $B^l$ is the total number of labeled samples. Once a cluster reaches its quota, we exclude further similar images to avoid semantic redundancy.

We set $B^u$ to 1M to construct the RS4P-1M dataset, which balances pseudo-label quality and semantic diversity, greatly improving model generalization and transferability.

\subsection{S4 Pre-training}

Through the above process, we construct the RS4P-1M dataset for S4P. We then introduce a general RSFMs pre-training framework that is compatible with existing S4 methods. Considering that more complex algorithms may suffer from low training efficiency on large-scale images, we adopt FixMatch \cite{fixmatch}, an efficient and widely used consistency regularization method that applies weak-to-strong data augmentations for pre-training. 

Consider a dataset consisting of labeled image pairs from the iSAID dataset \(\{(x^{l}_{i}, y^{l}_{i})\}_{i=1}^{B_l}\) and unlabeled images from our curated RS4P-1M collection \(\{x_j^{u}\}_{j=1}^{B_u}\). Each labeled image \(x_i^{l} \in \mathbb{R}^{H \times W \times 3}\) has corresponding pixel-level annotations \(y_i^{l} \in \mathbb{R}^{H \times W \times K}\), where \(K\) is the number of classes in the pre-training phase. Each unlabeled image \(x_j^{u} \in \mathbb{R}^{H \times W \times 3}\) has no annotations. The numbers of labeled and unlabeled images are \(B_l\) and \(B_u\), respectively.

FixMatch employs distinct transformation strategies for data augmentation: \textit{Weak augmentation \(\mathcal{A}_w\)} involves random scaling, cropping, rotation, and flipping, while \textit{Strong augmentation \(\mathcal{A}_s\)} applies more aggressive transformations, such as CutMix \cite{cutmix}, color jitter, grayscale conversion, and Gaussian blur.

For each unlabeled image \(x_j^u\), we generate two augmented views using sequential transformations:  
\begin{equation}  
x_j^{u_w} = \mathcal{A}_w(x_j^u), \quad x_j^{u_s} = \mathcal{A}_s(\mathcal{A}_w(x_j^u)).
\end{equation}  

The overall training objective function for both labeled and unlabeled images is given by:
\begin{equation}
\mathcal{L} = \mathcal{L}_s + \lambda \, \mathcal{L}_{u_s}.
\end{equation}

Here, the supervised and unsupervised loss terms are denoted as \(\mathcal{L}_s\) and \(\mathcal{L}_{u_s}\), respectively, with \(\lambda\) as a hyperparameter that controls the weight of the unsupervised loss. These losses are defined as:
\begin{equation}
\mathcal{L}_s = \frac{1}{B_l} \sum_{i=1}^{B_l} \mathcal{L}_{ce}(y^l_i, p_i),
\end{equation}
\begin{equation}
\mathcal{L}_{u_s} = \frac{1}{B_u} \sum_{j=1}^{B_u} \mathds{1}(\max(p_{j}^{u_w}) \geq \tau) \mathcal{L}_{ce}(\hat{y}_j, p_{j}^{u_s}),
\end{equation}
where \(\mathcal{L}_{ce}\) denotes the cross-entropy loss, \(p_i = f(x^{l}_{i})\), \(p_{j}^{u_w} = f(x_{j}^{u_w})\), and \(p_{j}^{u_s} = f(x_{j}^{u_s})\) are the predicted probability maps for labeled and unlabeled images. The pseudo-label \(\hat{y}_j\) is determined by \(\hat{y}_j = \arg\max(p_{j}^{u_w})\), with \(\tau\) being the confidence threshold for the pseudo-labels. The indicator function \(\mathds{1}\) ensures that only high-confidence predictions contribute to the unsupervised loss. During semi-supervised learning, the segmentation network \(f(\cdot)\) is optimized across both supervised and unsupervised branches, allowing it to learn more representative and discriminative features.

\begin{figure}[t] 
\centering
\includegraphics[width=0.4\textwidth]{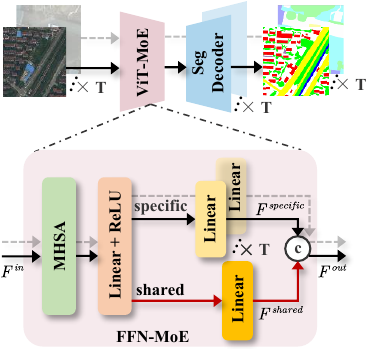}
\caption{The workflow of MoE-MDF. ViT-MoE refers to the incorporation of the FFN-MoE module into the standard ViT. The black solid arrows and dashed arrows represent the forward propagation paths for different datasets, while the red arrows indicate the shared forward propagation.}
\label{fig:FFN-MoE}
\end{figure}

\subsection{MoE-based Multiple Dataset Fine-tuning}

During fine-tuning, existing RSFMs often follow a "one dataset, one model" paradigm, requiring a separate model to be trained for each downstream dataset. This approach leads to significant parameter redundancy and lacks generalization across datasets, making it inefficient for unified deployment and scalability.

To address these issues, our goal is to develop general-purpose RSFMs that support multiple datasets for the same task (e.g., land cover segmentation), enabling unified fine-tuning and deployment. As shown in Figure \ref{fig:FFN-MoE}, we first adopt a shared backbone combined with dataset-specific decoders, and train the model via multiple-dataset joint fine-tuning (MDF) to handle $T$ datasets simultaneously. However, due to the considerable differences in data distributions, label spaces, and annotation styles across datasets, naively sharing the backbone may lead to feature interference, thereby hindering performance and convergence. To mitigate this, we further propose the MoE-MDF approach, which decouples shared and dataset-specific knowledge within the model. Specifically, we integrate the MoE modules into the ViT by splitting the FFN into multiple branches: a shared expert for learning general representations, and $T$ dataset-specific experts for modeling dataset-specific features. This design results in the FFN-MoE, as illustrated in Figure~\ref{fig:FFN-MoE}.

Taking a standard Transformer block as an example, the computation of FFN-MoE begins with the input feature $F^{\text{in}}$, it first passes through multi-head self-attention (MHSA), followed by a shared linear layer and a ReLU activation to obtain the intermediate feature:

\begin{equation}
F_{\text{FFN}} = \text{ReLU}(\text{Linear}(\text{MHSA}(F^{\text{in}})),
\end{equation} 
$F_{\text{FFN}} \in \mathbb{R}^{N \times D}$, where $N$ is the number of tokens and $D$ is the intermediate dimensionality. Next, this intermediate representation is passed through two parallel branches: one shared expert independent of the task, and one task-specific expert. The computation is defined as:

\begin{equation}
\begin{aligned}
F^{\text{shared}} &= \text{Linear}_{\text{shared}}^{ D \rightarrow (1 - \alpha)C}(F^{\text{FFN}}), \\
F^{\text{specific}} &= \text{Linear}_{\text{specific}}^{D \rightarrow \alpha C}(F^{\text{FFN}}),
\end{aligned}
\end{equation}
where $C$ is the output channel dimension, and $\alpha \in [0, 1]$ is a hyperparameter controlling the partition ratio between shared and task-specific capacities. The shared linear layer is updated using data from all tasks, while $\text{Linear}_{\text{specific}}^{(t)}$ is exclusively trained with samples from task $t$. Finally, the two outputs are concatenated along the channel dimension to form the final output of the FFN:

\begin{equation}
F^{\text{out}} = \text{Concat}\left(F^{\text{shared}},\ F^{\text{specific}} \right).
\end{equation}

Overall, the proposed MoE-MDF approach enables efficient and unified processing of multiple datasets, greatly reducing memory use and computational cost compared to maintaining separate models. It adds no extra parameters or inference delay, making it an efficient and practical universal backbone for RS segmentation. This method can also be extended to object detection, allowing one model to generalize across multiple datasets using the same fine-tuning and deployment pipeline. Overall, this unified framework provides a solid foundation for building general-purpose RSFMs that handle diverse downstream tasks effectively. 
\section{Experiments}
In this section, we comprehensively evaluate the proposed method on RS semantic segmentation and object detection tasks. S5 successfully pre-trains ViT RSFMs ranging from ViT-B to ViT-H (up to 600 million parameters) and achieves state-of-the-art performance on multiple RS benchmarks with fewer parameters after fine-tuning. We also validate the effectiveness of S5’s core components through ablation studies. More details of dataset, pre-training, and fine-tuning implementations details can be found in the appendix.

\begin{table*}[t]
  \centering
  \caption{
Comparison with existing RSFMs across multiple RS benchmarks. The evaluation metric for object detection is mAP, while for semantic segmentation it is mIoU, except for Potsdam, which uses mF1. Params Det (M) and Params Seg (M) indicate the number of parameters used for detection and segmentation models, respectively. Single and Multiple refer to the parameters required for handling a single dataset and multiple datasets.}
  \small
  \fontsize{8}{10.5}\selectfont
  \setlength{\tabcolsep}{5pt}
  \resizebox{\linewidth}{!}{
  \begin{tabular}{c|c|cc|cc|cc|cccc}
  \hline
  \multirow{3}{*}{Method} 
  & \multirow{3}{*}{Backbone} 
  & \multicolumn{2}{c|}{Params Det (M)} 
  & \multicolumn{2}{c|}{\textbf{Object Detection}} 
  & \multicolumn{2}{c|}{Params Seg (M)} 
  & \multicolumn{4}{c}{\textbf{Semantic Segmentation}} \\
  & & Single & Multiple & DIOR-R & DOTA-v2 & Single & Multiple & Vaihingen & Potsdam & LoveDA & OpenEarthMap \\
  \hline
  RVSA \cite{RVSA}         & ViT-B + RVSA   & 111.2 & 222.4  & 68.06 & 55.22 & 103.2 & 412.8 & 78.49 & 91.58 & 52.44 & 66.63 \\
  GFM \cite{GFM}           & Swin-B     & 104.1 & 208.2  & 67.67 & 59.15     & 96.9 & 387.6 & 79.61 & 91.85 & 54.98 & 67.78 \\
  Scale-MAE \cite{Scale-MAE} & ViT-L        & 334.6 & 669.2  & 66.47 & 56.97 & 327.4 & 1309.6 & 78.64 & 91.54 & 53.67 & 68.54 \\
  SAMRS \cite{SAMRS}       & ViT-B + RVSA   & - & -      & -     & -     & 103.2  & 412.8 & 78.73 & 91.69 & 53.04 & 67.37 \\
  SatMAE++ \cite{SatMAE++} & ViT-L          & 334.6 & 669.2  & 66.82 & 55.60 & 327.4 & 1309.6 & 78.80 & 91.64 & 52.82 & 65.62 \\
  BillionFM \cite{BillionFM} & ViT-G        & 996.9 & 1993.9 & 73.62 & 58.69 & 990.9 & - & -     & 92.58 & 54.40 & -     \\
  OREOLE \cite{OREOLE}     & ViT-G          & 996.9 & - & 71.31 & -     & 990.9 & - & -     & 92.20 & 54.00 & -     \\
  MTP \cite{MTP}           & ViT-L + RVSA   & 334.6 & 669.2  & 74.54 & 58.41 & 327.4 & 1309.6 & 80.62 & 92.47 & 54.16 & 69.04 \\
  MA3E \cite{MA3E}         & ViT-B          & 111.2 & - & 71.82 & -     & 103.2 & - & -  & 91.50 & -     & - \\
  SelectiveMAE \cite{SelectiveMAE} & ViT-L & 334.6 & 669.2  & 71.75 & 57.84     & 327.4 & 1309.6 & 80.45 & 92.78     & 54.31 & 69.30 \\
  \hline
  S5                      & ViT-B           & 111.2 & 138.3  & 72.95 & 57.20 & 103.2 & 160.4  & 79.85 & 92.40 & 54.02 & 68.65 \\
  S5                      & ViT-L           & 334.6  & 377.8  & 75.21 & 59.71 & 327.4 & 435.0  & 80.72 & 92.78 & \textbf{55.67} & 69.66 \\
  S5                      & ViT-H           & 671.7 & 730.0  & \textbf{75.30} & \textbf{59.89} & 663.4 & 824.5  & \textbf{80.85} & \textbf{92.97} & 55.65 & \textbf{70.02} \\
  \hline
  \end{tabular}
  }
  \label{tab:unified_rs_results}
\end{table*}

\subsection{Comparison with SOTA Methods}

To thoroughly assess our method, we fine-tune it on diverse RS benchmarks covering semantic segmentation and object detection. We compare S5 with state-of-the-art RSFMs including RVSA~\cite{RVSA}, GFM~\cite{GFM}, Scale-MAE~\cite{Scale-MAE}, SAMRS~\cite{SAMRS}, SatMAE++\cite{SatMAE++}, MTP\cite{MTP}, MA3E\cite{MA3E}, BillionFM~\cite{BillionFM}, OREOLE~\cite{OREOLE}, and SelectiveMAE~\cite{SelectiveMAE}. As summarized in Table~\ref{tab:unified_rs_results}, the results clearly demonstrate the superiority of S5 across a wide range of datasets.

\textbf{Object Detection.} We evaluate the detection performance of the proposed method on two representative RS object detection datasets: DIOR-R and DOTA-v2. As shown in Table \ref{tab:unified_rs_results}, under the Oriented R-CNN \cite{ORCN} detection framework, S5 consistently delivers superior performance across various backbone scales. From ViT-B, ViT-L to the larger ViT-H, S5 outperforms SOTA methods with comparable parameter sizes, and even matches or surpasses larger models while using fewer parameters. For example, with the ViT-L backbone, S5 achieves better performance than methods like MTP and SelectiveMAE, while requiring only about half the number of parameters when handling multiple datasets. These results demonstrate the strong scalability and generalization ability of S5 in RS object detection tasks.

\textbf{Semantic Segmentation.} We further validate the generalization capability of S5 on four challenging RS semantic segmentation datasets: Vaihingen, Potsdam, LoveDA, and OpenEarthMap. Built upon the UperNet \cite{UperNet} segmentation model, S5 brings consistent and significant performance improvements across all benchmarks. With the ViT-B backbone, S5 already outperforms methods such as RVSA and SAMRS with similar parameter sizes. As the backbone scales up to ViT-L and ViT-H, S5 continues to set new SOTA results across multiple datasets. Moreover, S5 demonstrates outstanding parameter efficiency in multi-dataset settings. 

For example, with the ViT-L backbone, it uses less than one-third of the segmentation parameters compared to Scale-MAE, SatMAE++, and SelectiveMAE, while delivering superior performance. This further highlights the scalability and generalization strength of S5 in RS semantic segmentation tasks.

\subsection{Ablation Studies}
In this section, we perform ablation studies to assess the effectiveness of each component in the S5 framework. Experiments are conducted on two semantic segmentation benchmarks (Vaihingen and LoveDA) and one object detection benchmark (DIOR-R). As LoveDA's test set requires online evaluation, all results on this dataset are reported on the official validation set.

\begin{table*}[t]
\centering
\caption{Comparison of S4P using different pre-training datasets.}
\small
  \fontsize{8}{10.5}\selectfont
  \setlength{\tabcolsep}{8.1pt}
   \resizebox{\linewidth}{!}{
\begin{tabular}{c|c|c|c|c|c|ccc}
\hline
\multirow{2}{*}{Labeled Data} & \multirow{2}{*}{Unlabeled Data} & \multirow{2}{*}{Images} & \multirow{2}{*}{Pre-training} & \multirow{2}{*}{Backbone} & S4 & \multicolumn{3}{c}{S4P (Pre-train $\rightarrow$ Finetune)} \\
& & & & & iSAID (Val) & Vaihingen & LoveDA & DIOR-R \\
\hline
\multirow{1}{*} - & - & - & MAE & ViT-B & 65.93 & 78.27 & 52.47 & 68.02 \\
\hline
\multirow{3}{*}{iSAID (Train)} 
& SAMRS & 100k & MAE + S4P & ViT-B & 67.59 & 79.61 & 53.66 & 69.13 \\
& MillionAID-random & 100k & MAE + S4P & ViT-B & 66.32 & 79.49 & 53.20 & 69.02 \\
& MillionAID* & 100k & MAE + S4P & ViT-B & \textbf{67.66} & \textbf{79.77} & \textbf{53.81} & \textbf{69.65} \\
\hline
\end{tabular}
}
\label{tab:dataset_curation}
\end{table*}

\begin{table*}[t!]
\centering
\caption{Comparison of different fine-tuning strategies. SDF (single‑dataset fine‑tuning): an independent segmentation model per dataset. MDF (multi‑dataset fine‑tuning): a shared backbone with dataset‑specific decoders.}
\small
\fontsize{8}{10.5}\selectfont
\setlength{\tabcolsep}{5pt}
\resizebox{\linewidth}{!}{
\begin{tabular}{c|c|c|c|c|c|cccc|c}
\hline
Pre-training & Backbone & Fine-tuning & Params (M) & GFLOPs & Resolution & Vaihingen & Potsdam & OpenEarthMap & LoveDA & Average \\
\hline
\multirow{1}{*}{MAE} 
& ViT-B & SDF & 412.8 & 178.3 & 512 $\times$ 512 & 78.27 & 91.58 & 66.23 & 52.47 & 72.14 \\
\hline
\multirow{5}{*}{MAE + S4P} 
& ViT-B & SDF & 412.8 & \multirow{5}{*}{178.3} & \multirow{5}{*}{512 $\times$ 512} & \textbf{79.93} & 92.24 & 67.35 & 54.51 & 73.51 \\
& ViT-B & MDF & 132.1 & & & 79.82 & 92.25 & 68.41 & 54.53 & 73.75 \\
& ViT-B-MoE ($\alpha=1/8$) & MoE-MDF & 146.2 & & & 79.76 & 92.30 & 68.52 & 54.62 & 73.80 \\
& ViT-B-MoE ($\alpha=1/4$) & MoE-MDF & 160.4 & & & 79.85 & \textbf{92.40} & \textbf{68.80} & 54.57 & \textbf{74.15} \\
& ViT-B-MoE ($\alpha=1/2$) & MoE-MDF & 188.7 & & & 79.84 & 92.39 & 68.66 & \textbf{54.64} & 73.88 \\
\hline
\end{tabular}
}
\label{tab:ffn_moe}
\end{table*}

\subsubsection{Effectiveness of pre-training Dataset Curation}

As shown in Table \ref{tab:dataset_curation}, we use the iSAID training set as the labeled data and take the MAE pre-trained ViT-B from the MillionAID dataset as our baseline. We then conduct S4P experiments with three different unlabeled datasets. We evaluate the generalization ability of S4P on three representative downstream benchmarks: semantic segmentation on Vaihingen and LoveDA, and object detection on DIOR-R. In addition, we assess performance on the iSAID validation set under the conventional S4 setting to evaluate the quality of pseudo-labels indirectly.

We first use the curated SAMRS dataset as the unlabeled data. The results show that introducing this large-scale RS images into S4P significantly improves performance on downstream segmentation and detection tasks. For example, we observe mIoU of 79.61 on Vaihingen, 53.66 on LoveDA and mAP of 69.13 on DIOR-R, both clearly surpassing the results obtained from MAE pre-training alone. This confirms that S4P serves as an effective complementary pre-training strategy for enhancing the performance of RSFMs on downstream tasks. Next, we randomly sample a subset of 100k images from MillionAID, denoted as MillionAID-random, to match the scale of SAMRS. Under this setting, the model shows a slight performance drop across all downstream tasks, with a notable decrease on LoveDA from 53.66 to 53.20. This suggests that a randomly sampled unlabeled set may suffer from data distribution mismatch and insufficient diversity, leading to lower pseudo-label quality and thus limiting the effectiveness of training. Meanwhile, the performance gain on the iSAID validation set is also less pronounced compared to SAMRS, rising only from 65.93 to 66.32, which further supports this observation.

To improve the reliability of pseudo-labels and the diversity of the unlabeled data, we propose a data selection strategy based on entropy filtering and diversity expansion, resulting in a curated subset named MillionAID*. Experimental results show that MillionAID* achieves the best performance across all evaluation tasks: 67.66 on the iSAID validation set, outperforming both SAMRS (67.59) and MillionAID-random (66.32); and 79.77, 53.81, and 69.65 on Vaihingen, LoveDA, and DIOR-R, respectively. These results clearly demonstrate that the proposed method effectively selects high-quality and diverse unlabeled images. Compared with the existing SAMRS dataset, our method offers better scalability, thereby further enhancing the generalization and transferability of S4P models.

\subsubsection{Scalability of S4P}

\begin{figure}[t] 
\centering
\includegraphics[width=0.49\textwidth]{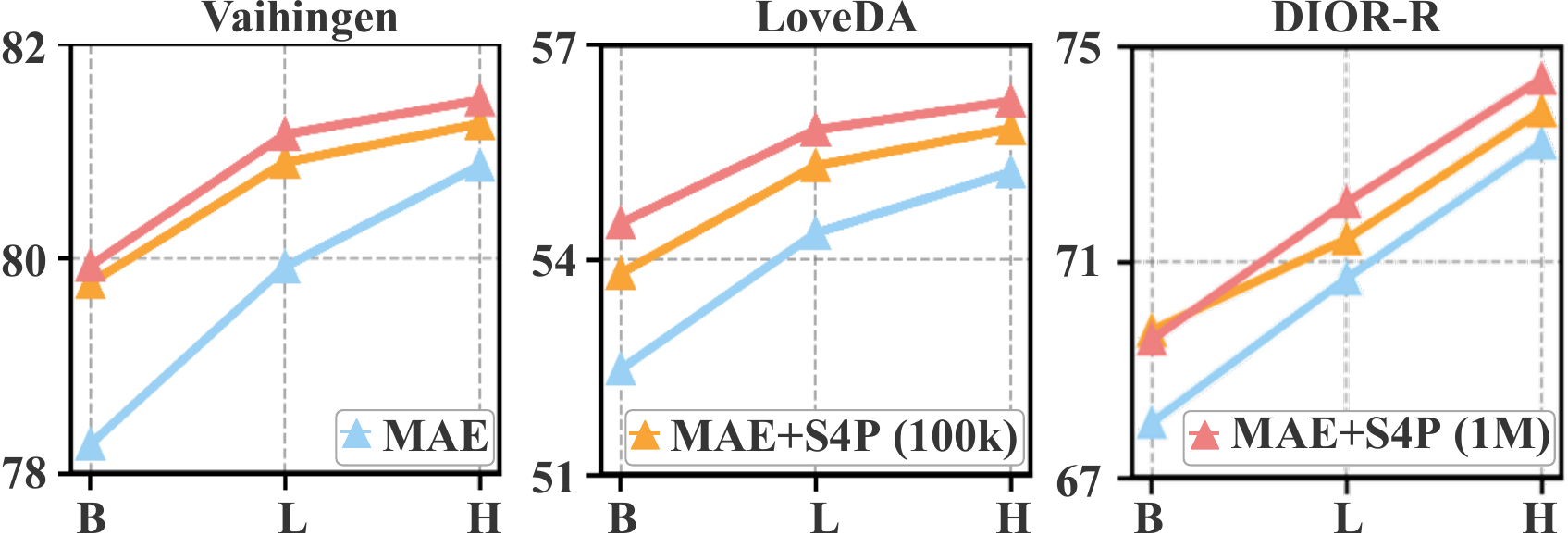}
\caption{Fine-tuning results on three RS benchmarks with varying pre-training dataset sizes and backbones. “100K” and “1M” indicate the number of images used for S4P.}
\label{fig:scalibility}
\end{figure}

We investigate the scalability of S4P with respect to both the model size and the scale of unlabeled data, as illustrated in Figure \ref{fig:scalibility}. Across all three downstream benchmarks, S4P consistently outperforms the MAE baseline, with performance further improving as model capacity or dataset scale increases. 
For instance, on the Vaihingen dataset, upgrading the backbone from ViT-B to ViT-H results in a significant performance boost for all pre-training setups. More importantly, expanding the unlabeled set from 100k to 1M images using our selection strategy leads to further improvements under each backbone. Similar trends are observed on the LoveDA and DIOR-R benchmarks. These results highlight two key scalability strengths of S4P: (1) effective utilization of large-scale unlabeled data when carefully selected for quality and diversity; and (2) consistently achieving higher performance with increasing model size. This underscores S4P’s potential as a general and scalable semi-supervised pre-training framework for RSFMs.

\subsubsection{The Ratio of Shared and Specific Experts}

Based on the results in Table \ref{tab:ffn_moe}, we evaluate various pre-training and fine-tuning strategies across four RS segmentation benchmarks, highlighting the advantages of MDF and the effect of the shared-to-specific expert ratio ($\alpha$) in the MoE-FFN design. The analysis is conducted using the average accuracy (Average) across the four datasets.

First, under the SDF paradigm, S4P effectively enhances the MAE backbone, improving performance across all four downstream tasks with the average accuracy rising from 72.14 to 73.51. Compared with SDF, MDF achieves consistently better results on all tasks, increasing the average accuracy to 73.75 while reducing parameters by nearly four times. This demonstrates that MDF enables the model to learn more general and robust feature representations under diverse data distributions. Finally, we introduce MoE-FFN to expand model capacity and analyze the impact of the specific expert ratio ($\alpha$). As $\alpha$ increases from 1/8 to 1/4, the average accuracy steadily improves from 73.80 to 74.15 across multiple datasets.

However, when $\alpha$ further increases to 1/2, performance gains plateau or slightly decline, likely because a high proportion of specific experts weakens the model’s generalization across datasets and reduces its adaptability to diverse data characteristics.

Balancing performance and parameter efficiency, we choose $\alpha=1/4$ as the optimal specific expert ratio in the MoE architecture. This setting delivers the best overall results across four benchmarks with only a slight increase in model size, demonstrating the value of a balanced design between shared and task-specific experts.
\section{Conclusion}
In this paper, we propose S5, a scalable semi-supervised semantic segmentation framework designed to pre-train RSFMs by leveraging vast unlabeled RS image data. S5 overcomes the limitations of prior S4 approaches constrained by small datasets and models, enabling large-scale semi-supervised pre-training. We introduce RS4P-1M, a million-scale curated dataset created via low-entropy filtering and diversity expansion, which serves as a solid basis for effective S4 pre-training of RSFMs with varying sizes. Additionally, S5 incorporates a MoE-MDF strategy to efficiently adapt RSFMs across multiple RS benchmarks with minimal parameter overhead. Extensive experiments demonstrate that RSFMs pre-trained under S5 achieve SOTA performance on diverse segmentation and detection benchmarks, highlighting the promise of scalable semi-supervised learning in advancing RS applications.
{
    \small
    \bibliographystyle{ieeenat_fullname}
    \bibliography{main}

\begin{thebibliography}{61}
\providecommand{\natexlab}[1]{#1}
\providecommand{\url}[1]{\texttt{#1}}
\expandafter\ifx\csname urlstyle\endcsname\relax
  \providecommand{\doi}[1]{doi: #1}\else
  \providecommand{\doi}{doi: \begingroup \urlstyle{rm}\Url}\fi

\bibitem[Cha et~al.(2023)Cha, Seo, and Lee]{BillionFM}
Keumgang Cha, Junghoon Seo, and Taekyung Lee.
\newblock A billion-scale foundation model for remote sensing images.
\newblock \emph{arXiv preprint arXiv:2304.05215}, 2023.

\bibitem[Chen and Shi(2020)]{LEVIR}
Hao Chen and Zhenwei Shi.
\newblock A spatial-temporal attention-based method and a new dataset for remote sensing image change detection.
\newblock \emph{Remote sensing}, 12\penalty0 (10):\penalty0 1662, 2020.

\bibitem[Cong et~al.(2022)Cong, Khanna, Meng, Liu, Rozi, He, Burke, Lobell, and Ermon]{SatMAE}
Yezhen Cong, Samar Khanna, Chenlin Meng, Patrick Liu, Erik Rozi, Yutong He, Marshall Burke, David Lobell, and Stefano Ermon.
\newblock Satmae: Pre-training transformers for temporal and multi-spectral satellite imagery.
\newblock \emph{Advances in Neural Information Processing Systems}, 35:\penalty0 197--211, 2022.

\bibitem[Cordts et~al.(2016)Cordts, Omran, Ramos, Rehfeld, Enzweiler, Benenson, Franke, Roth, and Schiele]{cityscapes}
Marius Cordts, Mohamed Omran, Sebastian Ramos, Timo Rehfeld, Markus Enzweiler, Rodrigo Benenson, Uwe Franke, Stefan Roth, and Bernt Schiele.
\newblock The cityscapes dataset for semantic urban scene understanding.
\newblock In \emph{In Proceedings of the IEEE/CVF International Conference on Computer Vision}, pages 3213--3223, 2016.

\bibitem[Deng et~al.(2009)Deng, Dong, Socher, Li, Li, and Fei-Fei]{ImageNet}
Jia Deng, Wei Dong, Richard Socher, Li-Jia Li, Kai Li, and Li Fei-Fei.
\newblock Imagenet: A large-scale hierarchical image database.
\newblock In \emph{In Proceedings of the IEEE/CVF Conference on Computer Vision and Pattern Recognition}, pages 248--255. Ieee, 2009.

\bibitem[Dias et~al.(2024)Dias, Tsaris, Bowman, Potnis, Arndt, Yang, and Lunga]{OREOLE}
Philipe Dias, Aristeidis Tsaris, Jordan Bowman, Abhishek Potnis, Jacob Arndt, H~Lexie Yang, and Dalton Lunga.
\newblock Oreole-fm: successes and challenges toward billion-parameter foundation models for high-resolution satellite imagery.
\newblock In \emph{Proceedings of the 32nd ACM International Conference on Advances in Geographic Information Systems}, pages 597--600, 2024.

\bibitem[Ding et~al.(2021)Ding, Xue, Xia, Bai, Yang, Yang, Belongie, Luo, Datcu, Pelillo, et~al.]{DOTA2}
Jian Ding, Nan Xue, Gui-Song Xia, Xiang Bai, Wen Yang, Michael~Ying Yang, Serge Belongie, Jiebo Luo, Mihai Datcu, Marcello Pelillo, et~al.
\newblock Object detection in aerial images: A large-scale benchmark and challenges.
\newblock \emph{IEEE Transactions on Pattern Analysis and Machine Intelligence}, 44\penalty0 (11):\penalty0 7778--7796, 2021.

\bibitem[Dosovitskiy et~al.(2020)Dosovitskiy, Beyer, Kolesnikov, Weissenborn, Zhai, Unterthiner, Dehghani, Minderer, Heigold, Gelly, et~al.]{ViT}
Alexey Dosovitskiy, Lucas Beyer, Alexander Kolesnikov, Dirk Weissenborn, Xiaohua Zhai, Thomas Unterthiner, Mostafa Dehghani, Matthias Minderer, Georg Heigold, Sylvain Gelly, et~al.
\newblock An image is worth 16x16 words: Transformers for image recognition at scale.
\newblock \emph{arXiv preprint arXiv:2010.11929}, 2020.

\bibitem[Everingham et~al.(2010)Everingham, Van~Gool, Williams, Winn, and Zisserman]{pascal}
Mark Everingham, Luc Van~Gool, Christopher~KI Williams, John Winn, and Andrew Zisserman.
\newblock The pascal visual object classes (voc) challenge.
\newblock \emph{International Journal of Computer Vision}, 88:\penalty0 303--338, 2010.

\bibitem[Fan et~al.(2023)Fan, Kukleva, Dai, and Schiele]{unimatch}
Yue Fan, Anna Kukleva, Dengxin Dai, and Bernt Schiele.
\newblock Revisiting consistency regularization for semi-supervised learning.
\newblock \emph{International Journal of Computer Vision}, 131\penalty0 (3):\penalty0 626--643, 2023.

\bibitem[French et~al.(2019)French, Laine, Aila, Mackiewicz, and Finlayson]{need}
Geoff French, Samuli Laine, Timo Aila, Michal Mackiewicz, and Graham Finlayson.
\newblock Semi-supervised semantic segmentation needs strong, varied perturbations.
\newblock \emph{arXiv preprint arXiv:1906.01916}, 2019.

\bibitem[Guo et~al.(2024)Guo, Lao, Dang, Zhang, Yu, Ru, Zhong, Huang, Wu, Hu, et~al.]{SkySense}
Xin Guo, Jiangwei Lao, Bo Dang, Yingying Zhang, Lei Yu, Lixiang Ru, Liheng Zhong, Ziyuan Huang, Kang Wu, Dingxiang Hu, et~al.
\newblock Skysense: A multi-modal remote sensing foundation model towards universal interpretation for earth observation imagery.
\newblock In \emph{In Proceedings of the IEEE/CVF Conference on Computer Vision and Pattern Recognition}, pages 27672--27683, 2024.

\bibitem[He et~al.(2022)He, Chen, Xie, Li, Doll{\'a}r, and Girshick]{MIM}
Kaiming He, Xinlei Chen, Saining Xie, Yanghao Li, Piotr Doll{\'a}r, and Ross Girshick.
\newblock Masked autoencoders are scalable vision learners.
\newblock In \emph{In Proceedings of the IEEE/CVF Conference on Computer Vision and Pattern Recognition}, pages 16000--16009, 2022.

\bibitem[Hoyer et~al.(2024)Hoyer, Tan, Naeem, Van~Gool, and Tombari]{SemiVL}
Lukas Hoyer, David~Joseph Tan, Muhammad~Ferjad Naeem, Luc Van~Gool, and Federico Tombari.
\newblock Semivl: semi-supervised semantic segmentation with vision-language guidance.
\newblock In \emph{In Proceedings of the European Conference on Computer Vision}, pages 257--275. Springer, 2024.

\bibitem[Huang et~al.(2024)Huang, Shi, Xiong, and Zhu]{DWL}
Wei Huang, Yilei Shi, Zhitong Xiong, and Xiao~Xiang Zhu.
\newblock Decouple and weight semi-supervised semantic segmentation of remote sensing images.
\newblock \emph{ISPRS Journal of Photogrammetry and Remote Sensing}, 212:\penalty0 13--26, 2024.

\bibitem[Jain et~al.(2022)Jain, Schoen-Phelan, and Ross]{multimodal}
Pallavi Jain, Bianca Schoen-Phelan, and Robert Ross.
\newblock Self-supervised learning for invariant representations from multi-spectral and sar images.
\newblock \emph{IEEE Journal of Selected Topics in Applied Earth Observations and Remote Sensing}, 15:\penalty0 7797--7808, 2022.

\bibitem[Ji et~al.(2018)Ji, Wei, and Lu]{WHU}
Shunping Ji, Shiqing Wei, and Meng Lu.
\newblock Fully convolutional networks for multisource building extraction from an open aerial and satellite imagery data set.
\newblock \emph{IEEE Transactions on Geoscience and Remote Sensing}, 57\penalty0 (1):\penalty0 574--586, 2018.

\bibitem[Kirillov et~al.(2023)Kirillov, Mintun, Ravi, Mao, Rolland, Gustafson, Xiao, Whitehead, Berg, Lo, et~al.]{SAM}
Alexander Kirillov, Eric Mintun, Nikhila Ravi, Hanzi Mao, Chloe Rolland, Laura Gustafson, Tete Xiao, Spencer Whitehead, Alexander~C Berg, Wan-Yen Lo, et~al.
\newblock Segment anything.
\newblock In \emph{In Proceedings of the IEEE/CVF International Conference on Computer Vision}, pages 4015--4026, 2023.

\bibitem[Li et~al.(2022)Li, Li, Zhang, Liu, Huang, Zhu, and Tao]{Global}
Haifeng Li, Yi Li, Guo Zhang, Ruoyun Liu, Haozhe Huang, Qing Zhu, and Chao Tao.
\newblock Global and local contrastive self-supervised learning for semantic segmentation of hr remote sensing images.
\newblock \emph{IEEE Transactions on Geoscience and Remote Sensing}, 60:\penalty0 1--14, 2022.

\bibitem[Li et~al.(2020)Li, Wan, Cheng, Meng, and Han]{DIOR}
Ke Li, Gang Wan, Gong Cheng, Liqiu Meng, and Junwei Han.
\newblock Object detection in optical remote sensing images: A survey and a new benchmark.
\newblock \emph{ISPRS journal of photogrammetry and remote sensing}, 159:\penalty0 296--307, 2020.

\bibitem[Li et~al.(2021)Li, Chen, Chen, and Shi]{Geographical}
Wenyuan Li, Keyan Chen, Hao Chen, and Zhenwei Shi.
\newblock Geographical knowledge-driven representation learning for remote sensing images.
\newblock \emph{IEEE Transactions on Geoscience and Remote Sensing}, 60:\penalty0 1--16, 2021.

\bibitem[Li et~al.(2024{\natexlab{a}})Li, Wang, Wang, Yang, Luo, Wang, Deng, Wang, Sun, Li, et~al.]{STAR}
Yansheng Li, Linlin Wang, Tingzhu Wang, Xue Yang, Junwei Luo, Qi Wang, Youming Deng, Wenbin Wang, Xian Sun, Haifeng Li, et~al.
\newblock Star: A first-ever dataset and a large-scale benchmark for scene graph generation in large-size satellite imagery.
\newblock \emph{IEEE Transactions on Pattern Analysis and Machine Intelligence}, 2024{\natexlab{a}}.

\bibitem[Li et~al.(2023)Li, Chen, Wu, Li, and Jing]{SegMind}
Zhenghong Li, Hao Chen, Jiangjiang Wu, Jun Li, and Ning Jing.
\newblock Segmind: Semisupervised remote sensing image semantic segmentation with masked image modeling and contrastive learning method.
\newblock \emph{IEEE Transactions on Geoscience and Remote Sensing}, 61:\penalty0 1--17, 2023.

\bibitem[Li et~al.(2024{\natexlab{b}})Li, Hou, Ma, Wu, Guo, Ren, and Jiao]{MA3E}
Zhihao Li, Biao Hou, Siteng Ma, Zitong Wu, Xianpeng Guo, Bo Ren, and Licheng Jiao.
\newblock Masked angle-aware autoencoder for remote sensing images.
\newblock In \emph{In Proceedings of the European Conference on Computer Vision}, pages 260--278. Springer, 2024{\natexlab{b}}.

\bibitem[Lin et~al.(2014)Lin, Maire, Belongie, Hays, Perona, Ramanan, Doll{\'a}r, and Zitnick]{COCO}
Tsung-Yi Lin, Michael Maire, Serge Belongie, James Hays, Pietro Perona, Deva Ramanan, Piotr Doll{\'a}r, and C~Lawrence Zitnick.
\newblock Microsoft coco: Common objects in context.
\newblock In \emph{In Proceedings of the European Conference on Computer Vision}, pages 740--755. Springer, 2014.

\bibitem[Long et~al.(2021)Long, Xia, Li, Yang, Yang, Zhu, Zhang, and Li]{MillionAID}
Yang Long, Gui-Song Xia, Shengyang Li, Wen Yang, Michael~Ying Yang, Xiao~Xiang Zhu, Liangpei Zhang, and Deren Li.
\newblock On creating benchmark dataset for aerial image interpretation: Reviews, guidances, and million-aid.
\newblock \emph{IEEE Journal of Selected Topics in Applied Earth Observations and Remote Sensing}, 14:\penalty0 4205--4230, 2021.

\bibitem[Lu et~al.(2023)Lu, Jiao, Li, Liu, Liu, Yang, Feng, and Chen]{WSCL}
Xiaoqiang Lu, Licheng Jiao, Lingling Li, Fang Liu, Xu Liu, Shuyuan Yang, Zhixi Feng, and Puhua Chen.
\newblock Weak-to-strong consistency learning for semisupervised image segmentation.
\newblock \emph{IEEE Transactions on Geoscience and Remote Sensing}, 61:\penalty0 1--15, 2023.

\bibitem[Mai et~al.(2024)Mai, Sun, Zhang, and Wu]{RankMatch}
Huayu Mai, Rui Sun, Tianzhu Zhang, and Feng Wu.
\newblock Rankmatch: Exploring the better consistency regularization for semi-supervised semantic segmentation.
\newblock In \emph{In Proceedings of the IEEE/CVF Conference on Computer Vision and Pattern Recognition}, pages 3391--3401, 2024.

\bibitem[Mall et~al.(2023)Mall, Hariharan, and Bala]{temporal}
Utkarsh Mall, Bharath Hariharan, and Kavita Bala.
\newblock Change-aware sampling and contrastive learning for satellite images.
\newblock In \emph{In Proceedings of the IEEE/CVF Conference on Computer Vision and Pattern Recognition}, pages 5261--5270, 2023.

\bibitem[Mendieta et~al.(2023)Mendieta, Han, Shi, Zhu, and Chen]{GFM}
Mat{\'\i}as Mendieta, Boran Han, Xingjian Shi, Yi Zhu, and Chen Chen.
\newblock Towards geospatial foundation models via continual pretraining.
\newblock In \emph{In Proceedings of the IEEE/CVF International Conference on Computer Vision}, pages 16806--16816, 2023.

\bibitem[Noman et~al.(2024)Noman, Naseer, Cholakkal, Anwer, Khan, and Khan]{SatMAE++}
Mubashir Noman, Muzammal Naseer, Hisham Cholakkal, Rao~Muhammad Anwer, Salman Khan, and Fahad~Shahbaz Khan.
\newblock Rethinking transformers pre-training for multi-spectral satellite imagery.
\newblock In \emph{In Proceedings of the IEEE/CVF Conference on Computer Vision and Pattern Recognition}, pages 27811--27819, 2024.

\bibitem[Oquab et~al.(2024)Oquab, Darcet, Moutakanni, Vo, Szafraniec, Khalidov, Fernandez, Haziza, Massa, El-Nouby, et~al.]{DINOV2}
Maxime Oquab, Timoth{\'e}e Darcet, Th{\'e}o Moutakanni, Huy Vo, Marc Szafraniec, Vasil Khalidov, Pierre Fernandez, Daniel Haziza, Francisco Massa, Alaaeldin El-Nouby, et~al.
\newblock Dinov2: Learning robust visual features without supervision.
\newblock \emph{Transactions on Machine Learning Research Journal}, pages 1--31, 2024.

\bibitem[Ouali et~al.(2020)Ouali, Hudelot, and Tami]{CCT}
Yassine Ouali, C{\'e}line Hudelot, and Myriam Tami.
\newblock Semi-supervised semantic segmentation with cross-consistency training.
\newblock In \emph{In Proceedings of the IEEE/CVF Conference on Computer Vision and Pattern Recognition}, pages 12674--12684, 2020.

\bibitem[Reed et~al.(2023)Reed, Gupta, Li, Brockman, Funk, Clipp, Keutzer, Candido, Uyttendaele, and Darrell]{Scale-MAE}
Colorado~J Reed, Ritwik Gupta, Shufan Li, Sarah Brockman, Christopher Funk, Brian Clipp, Kurt Keutzer, Salvatore Candido, Matt Uyttendaele, and Trevor Darrell.
\newblock Scale-mae: A scale-aware masked autoencoder for multiscale geospatial representation learning.
\newblock In \emph{In Proceedings of the IEEE/CVF International Conference on Computer Vision}, pages 4088--4099, 2023.

\bibitem[Sohn et~al.(2020)Sohn, Berthelot, Carlini, Zhang, Zhang, Raffel, Cubuk, Kurakin, and Li]{fixmatch}
Kihyuk Sohn, David Berthelot, Nicholas Carlini, Zizhao Zhang, Han Zhang, Colin~A Raffel, Ekin~Dogus Cubuk, Alexey Kurakin, and Chun-Liang Li.
\newblock Fixmatch: Simplifying semi-supervised learning with consistency and confidence.
\newblock \emph{Advances in Neural Information Processing Systems}, 33:\penalty0 596--608, 2020.

\bibitem[Souly et~al.(2017)Souly, Spampinato, and Shah]{semiGAN}
Nasim Souly, Concetto Spampinato, and Mubarak Shah.
\newblock Semi supervised semantic segmentation using generative adversarial network.
\newblock In \emph{In Proceedings of the IEEE/CVF International Conference on Computer Vision}, pages 5688--5696, 2017.

\bibitem[Stojnic and Risojevic(2021)]{multiview}
Vladan Stojnic and Vladimir Risojevic.
\newblock Self-supervised learning of remote sensing scene representations using contrastive multiview coding.
\newblock In \emph{In Proceedings of the IEEE/CVF Conference on Computer Vision and Pattern Recognition}, pages 1182--1191, 2021.

\bibitem[Sun et~al.(2024)Sun, Yang, Zhang, Cheng, and Hou]{CorrMatch}
Boyuan Sun, Yuqi Yang, Le Zhang, Ming-Ming Cheng, and Qibin Hou.
\newblock Corrmatch: Label propagation via correlation matching for semi-supervised semantic segmentation.
\newblock In \emph{In Proceedings of the IEEE/CVF Conference on Computer Vision and Pattern Recognition}, pages 3097--3107, 2024.

\bibitem[Sun et~al.(2022)Sun, Wang, Lu, Zhu, Lu, He, Li, Rong, Yang, Chang, et~al.]{RingMo}
Xian Sun, Peijin Wang, Wanxuan Lu, Zicong Zhu, Xiaonan Lu, Qibin He, Junxi Li, Xuee Rong, Zhujun Yang, Hao Chang, et~al.
\newblock Ringmo: A remote sensing foundation model with masked image modeling.
\newblock \emph{IEEE Transactions on Geoscience and Remote Sensing}, 61:\penalty0 1--22, 2022.

\bibitem[Tian et~al.(2020)Tian, Sun, Poole, Krishnan, Schmid, and Isola]{contrastive}
Yonglong Tian, Chen Sun, Ben Poole, Dilip Krishnan, Cordelia Schmid, and Phillip Isola.
\newblock What makes for good views for contrastive learning?
\newblock \emph{Advances in Neural Information Processing Systems}, 33:\penalty0 6827--6839, 2020.

\bibitem[Wang et~al.(2022{\natexlab{a}})Wang, Zhang, Du, Xia, and Tao]{RSP}
Di Wang, Jing Zhang, Bo Du, Gui-Song Xia, and Dacheng Tao.
\newblock An empirical study of remote sensing pretraining.
\newblock \emph{IEEE Transactions on Geoscience and Remote Sensing}, 61:\penalty0 1--20, 2022{\natexlab{a}}.

\bibitem[Wang et~al.(2022{\natexlab{b}})Wang, Zhang, Xu, Zhang, Du, Tao, and Zhang]{RVSA}
Di Wang, Qiming Zhang, Yufei Xu, Jing Zhang, Bo Du, Dacheng Tao, and Liangpei Zhang.
\newblock Advancing plain vision transformer toward remote sensing foundation model.
\newblock \emph{IEEE Transactions on Geoscience and Remote Sensing}, 61:\penalty0 1--15, 2022{\natexlab{b}}.

\bibitem[Wang et~al.(2023)Wang, Zhang, Du, Xu, Liu, Tao, and Zhang]{SAMRS}
Di Wang, Jing Zhang, Bo Du, Minqiang Xu, Lin Liu, Dacheng Tao, and Liangpei Zhang.
\newblock Samrs: Scaling-up remote sensing segmentation dataset with segment anything model.
\newblock \emph{Advances in Neural Information Processing Systems}, 36:\penalty0 8815--8827, 2023.

\bibitem[Wang et~al.(2024{\natexlab{a}})Wang, Zhang, Xu, Liu, Wang, Gao, Han, Guo, Du, Tao, et~al.]{MTP}
Di Wang, Jing Zhang, Minqiang Xu, Lin Liu, Dongsheng Wang, Erzhong Gao, Chengxi Han, Haonan Guo, Bo Du, Dacheng Tao, et~al.
\newblock Mtp: Advancing remote sensing foundation model via multi-task pretraining.
\newblock \emph{IEEE Journal of Selected Topics in Applied Earth Observations and Remote Sensing}, 2024{\natexlab{a}}.

\bibitem[Wang et~al.(2025)Wang, Wang, Wang, Guo, Zhong, Lan, Yang, and Zhang]{SelectiveMAE}
Fengxiang Wang, Hongzhen Wang, Di Wang, Zonghao Guo, Zhenyu Zhong, Long Lan, Wenjing Yang, and Jing Zhang.
\newblock Harnessing massive satellite imagery with efficient masked image modeling.
\newblock \emph{arXiv preprint arXiv:2406.11933}, 2025.

\bibitem[Wang et~al.(2024{\natexlab{b}})Wang, Zhang, Li, and Li]{wang2024allspark}
Haonan Wang, Qixiang Zhang, Yi Li, and Xiaomeng Li.
\newblock Allspark: Reborn labeled features from unlabeled in transformer for semi-supervised semantic segmentation.
\newblock In \emph{In Proceedings of the IEEE/CVF Conference on Computer Vision and Pattern Recognition}, pages 3627--3636, 2024{\natexlab{b}}.

\bibitem[Wang et~al.(2021{\natexlab{a}})Wang, Zheng, Ma, Lu, and Zhong]{LoveDA}
Junjue Wang, Zhuo Zheng, Ailong Ma, Xiaoyan Lu, and Yanfei Zhong.
\newblock Loveda: A remote sensing land-cover dataset for domain adaptive semantic segmentation.
\newblock In \emph{Advances in Neural Information Processing Systems}, 2021{\natexlab{a}}.

\bibitem[Wang et~al.(2021{\natexlab{b}})Wang, Chen, Ding, Tang, and Luo]{RanPaste}
Jia-Xin Wang, Si-Bao Chen, Chris~HQ Ding, Jin Tang, and Bin Luo.
\newblock Ranpaste: Paste consistency and pseudo label for semisupervised remote sensing image semantic segmentation.
\newblock \emph{IEEE Transactions on Geoscience and Remote Sensing}, 60:\penalty0 1--16, 2021{\natexlab{b}}.

\bibitem[Waqas~Zamir et~al.(2019)Waqas~Zamir, Arora, Gupta, Khan, Sun, Shahbaz~Khan, Zhu, Shao, Xia, and Bai]{iSAID}
Syed Waqas~Zamir, Aditya Arora, Akshita Gupta, Salman Khan, Guolei Sun, Fahad Shahbaz~Khan, Fan Zhu, Ling Shao, Gui-Song Xia, and Xiang Bai.
\newblock isaid: A large-scale dataset for instance segmentation in aerial images.
\newblock In \emph{In Proceedings of the IEEE/CVF Conference on Computer Vision and Pattern Recognition Workshops}, pages 28--37, 2019.

\bibitem[Xia et~al.(2023)Xia, Yokoya, Adriano, and Broni-Bediako]{Openearthmap}
Junshi Xia, Naoto Yokoya, Bruno Adriano, and Clifford Broni-Bediako.
\newblock Openearthmap: A benchmark dataset for global high-resolution land cover mapping.
\newblock In \emph{Proceedings of the IEEE/CVF Winter Conference on Applications of Computer Vision}, pages 6254--6264, 2023.

\bibitem[Xiao et~al.(2018)Xiao, Liu, Zhou, Jiang, and Sun]{UperNet}
Tete Xiao, Yingcheng Liu, Bolei Zhou, Yuning Jiang, and Jian Sun.
\newblock Unified perceptual parsing for scene understanding.
\newblock In \emph{In Proceedings of the European Conference on Computer Vision}, pages 418--434, 2018.

\bibitem[Xie et~al.(2021)Xie, Cheng, Wang, Yao, and Han]{ORCN}
Xingxing Xie, Gong Cheng, Jiabao Wang, Xiwen Yao, and Junwei Han.
\newblock Oriented r-cnn for object detection.
\newblock In \emph{In Proceedings of the IEEE/CVF International Conference on Computer Vision}, pages 3520--3529, 2021.

\bibitem[Yang et~al.(2022)Yang, Zhuo, Qi, Shi, and Gao]{ST++}
Lihe Yang, Wei Zhuo, Lei Qi, Yinghuan Shi, and Yang Gao.
\newblock St++: Make self-training work better for semi-supervised semantic segmentation.
\newblock In \emph{In Proceedings of the IEEE/CVF Conference on Computer Vision and Pattern Recognition}, pages 4268--4277, 2022.

\bibitem[Yang et~al.(2025)Yang, Zhao, and Zhao]{UnimatchV2}
Lihe Yang, Zhen Zhao, and Hengshuang Zhao.
\newblock Unimatch v2: Pushing the limit of semi-supervised semantic segmentation.
\newblock \emph{IEEE Transactions on Pattern Analysis and Machine Intelligence}, 2025.

\bibitem[Yun et~al.(2019)Yun, Han, Oh, Chun, Choe, and Yoo]{cutmix}
Sangdoo Yun, Dongyoon Han, Seong~Joon Oh, Sanghyuk Chun, Junsuk Choe, and Youngjoon Yoo.
\newblock Cutmix: Regularization strategy to train strong classifiers with localizable features.
\newblock In \emph{In Proceedings of the IEEE/CVF International Conference on Computer Vision}, pages 6023--6032, 2019.

\bibitem[Zhang et~al.(2025)Zhang, Chen, Liu, Chen, Zou, and Shi]{CDMAMBA}
Haotian Zhang, Keyan Chen, Chenyang Liu, Hao Chen, Zhengxia Zou, and Zhenwei Shi.
\newblock Cdmamba: Incorporating local clues into mamba for remote sensing image binary change detection.
\newblock \emph{IEEE Transactions on Geoscience and Remote Sensing}, 2025.

\bibitem[Zhang and Zhang(2022)]{zhang2022artificial}
Lefei Zhang and Liangpei Zhang.
\newblock Artificial intelligence for remote sensing data analysis: A review of challenges and opportunities.
\newblock \emph{IEEE Geoscience and Remote Sensing Magazine}, 10\penalty0 (2):\penalty0 270--294, 2022.

\bibitem[Zhao et~al.(2023{\natexlab{a}})Zhao, Long, Pi, Wang, and Zhou]{iMAS}
Zhen Zhao, Sifan Long, Jimin Pi, Jingdong Wang, and Luping Zhou.
\newblock Instance-specific and model-adaptive supervision for semi-supervised semantic segmentation.
\newblock In \emph{In Proceedings of the IEEE/CVF Conference on Computer Vision and Pattern Recognition}, pages 23705--23714, 2023{\natexlab{a}}.

\bibitem[Zhao et~al.(2023{\natexlab{b}})Zhao, Yang, Long, Pi, Zhou, and Wang]{AugSeg}
Zhen Zhao, Lihe Yang, Sifan Long, Jimin Pi, Luping Zhou, and Jingdong Wang.
\newblock Augmentation matters: A simple-yet-effective approach to semi-supervised semantic segmentation.
\newblock In \emph{In Proceedings of the IEEE/CVF Conference on Computer Vision and Pattern Recognition}, pages 11350--11359, 2023{\natexlab{b}}.

\bibitem[Zhou et~al.(2017)Zhou, Zhao, Puig, Fidler, Barriuso, and Torralba]{ade20k}
Bolei Zhou, Hang Zhao, Xavier Puig, Sanja Fidler, Adela Barriuso, and Antonio Torralba.
\newblock Scene parsing through ade20k dataset.
\newblock In \emph{In Proceedings of the IEEE/CVF International Conference on Computer Vision}, pages 633--641, 2017.

\bibitem[Zhu et~al.(2021)Zhu, Cao, Zhai, Cheng, and Zha]{Self-promoted}
Kai Zhu, Yang Cao, Wei Zhai, Jie Cheng, and Zheng-Jun Zha.
\newblock Self-promoted prototype refinement for few-shot class-incremental learning.
\newblock In \emph{In Proceedings of the IEEE/CVF Conference on Computer Vision and Pattern Recognition}, pages 6801--6810, 2021.

\end{thebibliography}
}

\clearpage
\setcounter{page}{1}
\maketitlesupplementary

\section{Pre-training Dataset and Implementations}

\begin{figure*}[t] 
\centering
\includegraphics[width=\textwidth]{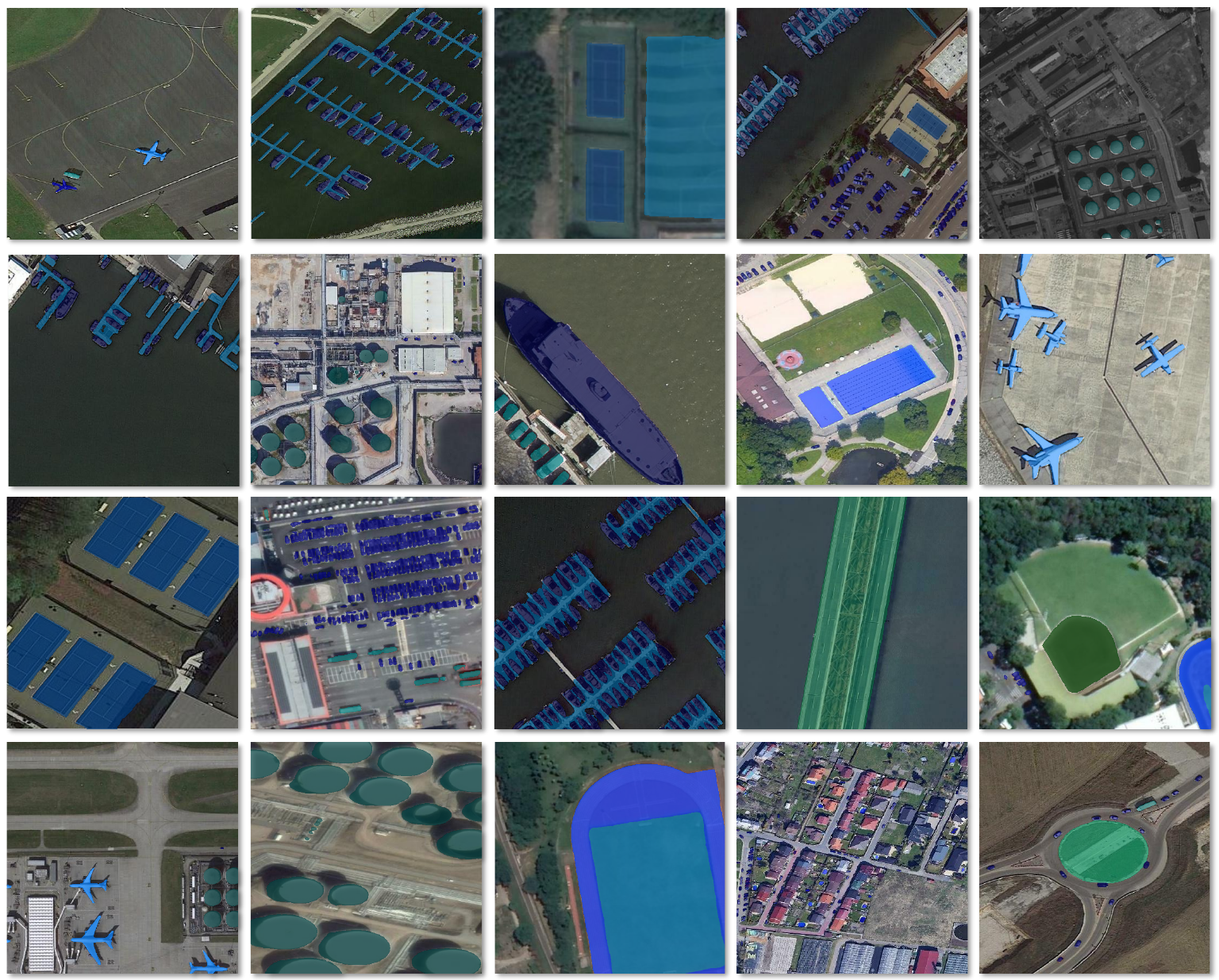}
\caption{
Visualization of RS4P-1M samples with generated pseudo-labels.
}
\label{fig:RS4P-1M}
\end{figure*}

\textbf{RS4P-1M.} We obtain one million unlabeled RS images by cropping and filtering samples from SAMRS \cite{SAMRS}, STAR \cite{STAR}, and MillionAID \cite{MillionAID}. To facilitate training, we follow standard practices in the RS community \cite{RSP} and crop large-scale images into uniform patches of 512×512 pixels. The number of semantic clusters $M$ in the clustering process is set to 150. Through careful processing of these datasets, we construct RS4P-1M, a large-scale S4P resource that covers diverse geospatial scenes and object types. Figure~\ref{fig:RS4P-1M} shows sample visualizations of RS4P-1M along with pseudo-labels generated using UperNet with a ViT-H backbone.

\noindent 
\textbf{Implementation Details.} All experiments are conducted using the PyTorch framework on 8 NVIDIA RTX 4090 GPUs. We adopt the widely used segmentation model UPerNet \cite{UperNet}, with ViT-B, ViT-L, and ViT-H \cite{ViT} as backbones. The backbones are initialized with MAE pre-trained weights on the MillionAID \cite{MillionAID} dataset. Models are trained using the AdamW optimizer for 120,000 iterations, with a weight decay of 0.01 and a cosine learning rate schedule. The base learning rate is set to 5e-5, with $\tau$ set to the default value of 0.95 and $\lambda$ set to 1.0. The batch size is 48 for ViT-H and 96 for the other models. To optimize memory usage and training efficiency, we adopt mixed-precision training and gradient checkpointing.

\section{Fine-tuning Datasets and Implementations}

\subsection{Semantic Segmentation}
\noindent \textbf{OpenEarthMap} \cite{Openearthmap}  is a benchmark dataset for global high-resolution land cover mapping. It features satellite and aerial images with a ground sampling distance between 0.25 and 0.5 meters. The dataset is manually annotated with nine semantic classes—background, bareland, rangeland, developed space, road, tree, water, agricultural land, and building—plus a background category. Its wide geographic coverage spans 97 regions across 44 countries on six continents, and for evaluation, only the validation subset (excluding xBD data) is used.

\noindent 
\textbf{LoveDA} \cite{LoveDA}  is designed for domain-adaptive semantic segmentation in RS. It consists of 5,987 high-resolution images (0.3 m) collected from both urban and rural areas in cities such as Nanjing, Changzhou, and Wuhan. The dataset labels seven categories: building, road, water, barren, forest, agriculture, and background.

\noindent 
\textbf{Potsdam}\footnote{http://www2.isprs.org/commissions/comm3/wg4/2d-sem-label-potsdam.html} and \textbf{Vaihingen} \footnote{http://www2.isprs.org/commissions/comm3/wg4/2d-sem-label-vaihingen.html} datasets are established benchmarks for urban semantic segmentation. Potsdam offers images at an ultra-high resolution of 5 cm, while Vaihingen provides 9 cm resolution imagery. Both datasets are annotated with six classes: typically, impervious surfaces (e.g., roads and parking lots), buildings, low vegetation, trees, cars, and clutter. In the experiments, we follow \cite{SAMRS} and ignore the clutter class.

\subsection{Object Detection}

\textbf{DIOR-R } \cite{DIOR} is a rotation-based extension of the DIOR dataset for object detection in RS images. It contains 23,463 optical RS images with approximately 192,518 object instances across 20 categories, including airplane, airport, bridge, ship, storage tank, stadium, and more. Unlike the original DIOR dataset, which uses horizontal bounding boxes, DIOR-R employs oriented bounding boxes (OBB) to more accurately capture the spatial layout and rotation of objects.

\noindent 
\textbf{DOTA-v2.0} \cite{DOTA2} is a large-scale benchmark for RS object detection. It consists of 11,268 high-resolution images across 18 categories. Compared to its earlier version, DOTA-v2.0 introduces new categories such as "airport" and "helipad". The images are sourced from Google Earth, GF-2 satellites, and various aerial platforms. Each object is annotated with an 8-degree-of-freedom oriented quadrilateral, enabling precise localization of multi-scale and arbitrarily rotated targets.

\subsection{Implementation Details} The fine-tuning experiments are conducted on 4 NVIDIA RTX 4090 GPUs and 4 NVIDIA L20 GPUs. We evaluate the proposed S5 method using ViT-B, ViT-L, and ViT-H backbones on four RS semantic segmentation benchmarks and two RS object detection datasets. Detailed training configurations are provided in Table~\ref{impletation}. For ease of comparison, all parameter counts are measured at an input resolution of $512 \times 512$.

\begin{table*}[ht]
\centering
\caption{Fine-tuning implementation details.}
\small
\fontsize{8}{10.5}\selectfont
\setlength{\tabcolsep}{8.1pt}
\resizebox{\textwidth}{!}{
\begin{tabular}{l|cccc|cc}
\hline
Task & \multicolumn{4}{c|}{Semantic Segmentation} & \multicolumn{2}{c}{Object Detection} \\
\hline
 Dataset & Vaihingen & Potsdam & LoveDA & OpenEarthMap & DIOR-R & DOTA-v2.0 \\
\hline
Optimizer & AdamW & AdamW & AdamW & AdamW & AdamW & AdamW \\
Input Size & 512 $\times$ 512 & 512 $\times$ 512 & 512 $\times$ 512 & 512 $\times$ 512 & 1024 $\times$ 1024 & 1024 $\times$ 1024 \\
Input channel & RGB & NIRRG & RGB & RGB & RGB & RGB \\
Base learning rate & 5e-5 & 5e-5 & 5e-5 & 5e-5 & 1e-4 & 1e-4 \\
Learning rate scheduler & Cosine Annealing & Cosine Annealing & Cosine Annealing & Cosine Annealing & Multistep & Multistep \\
Weight decay & 0.05 & 0.05 & 0.05 & 0.05 & 0.05 & 0.05 \\
Batch size & 24 & 24 & 24 & 24 & 8 & 8 \\
Max iteration/epoch & 100 epoch & 100 epoch & 100 epoch & 100 epoch & 12 epoch & 12 epoch \\
Augmentation &
\makecell[c]{RandomScaling \\ (0.5 to 2.0)\\ RandomCrop, \\ RandomFlip, \\
RandomRotate,\\ ColorJitter} &
\makecell[c]{RandomScaling \\ (0.5 to 2.0), \\ RandomCrop, \\ RandomFlip, \\
RandomRotate,\\ ColorJitter} &
\makecell[c]{RandomScaling \\ (0.5 to 2.0), \\ RandomCrop, \\ RandomFlip, \\
RandomRotate,\\ ColorJitter}&
\makecell[c]{RandomScaling \\ (0.5 to 2.0), \\
RandomCrop} &
\makecell[c]{RandomFlip} &
\makecell[c]{RandomFlip} \\
Head/Detector & UPerNet & UPerNet & UPerNet & UPerNet & Oriented-RCNN & Oriented-RCNN \\
\hline
\end{tabular}
}
\label{impletation}
\end{table*}

\section{Additional Experiments}

\subsection{Fine-Tuning Evaluation on Change Detection}
We also evaluate the transferability of S4P-pretrained backbones on change detection tasks, validating performance on the WHU \cite{WHU} and LEVIR-CD \cite{LEVIR} datasets. All fine-tuning settings follow the protocols of MTP \cite{MTP} and CDMamba \cite{CDMAMBA}. As shown in Table~\ref{change_detection}, S4P significantly outperforms MAE on both benchmarks, further demonstrating its strong generalization capability.

\begin{table}[t]
\centering
\caption{Comparison of fine-tuning results (F1 score) using different pre-training strategies on change detection benchmarks (WHU and LEVIR-CD).}
\small
\fontsize{8}{10.5}\selectfont
\setlength{\tabcolsep}{8.1pt}
\resizebox{\linewidth}{!}{
\begin{tabular}{llcc}
\hline
Method  & Backbone & WHU & LEVIR-CD \\
\hline   
 MAE & ViT-B   & 94.39 & 91.92 \\
 MAE + S4P  & ViT-B   & \textbf{94.97} & \textbf{92.14} \\
\hline
MAE & ViT-L   & 94.92 & 92.26 \\
 MAE + S4P  & ViT-L   & \textbf{95.26} & \textbf{92.37} \\
\hline
 MAE & ViT-H   & 95.36 & 92.70 \\
 MAE + S4P  & ViT-H   & \textbf{95.66} & \textbf{92.75} \\
\hline
\end{tabular}}
\label{change_detection}
\end{table}

\begin{table*}[t]
  \centering
  \caption{Comparison of  S4P via different S4 methods in terms of fine-tuning performance on the Vaihingen, LoveDA, and DIOR-R datasets. GPU memory usage and pre-training time are measured on an NVIDIA RTX 4090 GPU with a batch size of 8.
  }
\small
\fontsize{8}{10.5}\selectfont
\setlength{\tabcolsep}{8.1pt}  
 \resizebox{0.8\linewidth}{!}{
  \begin{tabular}{c|ccccc}
     \hline
    Method  &  Memory (G) / GPU & Time (h) / epoch &  Vaihingen  & LoveDA & DIOR-R\\
        \hline
    FixMatch & 9.51  & 7.10  &  79.45 & 53.85 & \textbf{69.33} \\ 
    UniMatch   & 13.44  & 8.34 &  79.43 & \textbf{54.10} & 69.22\\ 
    CorrMatch \ & 20.45  & 11.12  &  \textbf{79.50} & 53.94 & 69.12 \\  
   \hline
  \end{tabular}
  }
  \label{tab:s4}
\end{table*}

\subsection{Comparison of Various S4 Methods}

We investigate the impact of different S4 pre-training methods on fine-tuning performance within the S5 framework. For a fair comparison, we follow the official implementations of UniMatch \cite{unimatch} and CorrMatch \cite{CorrMatch}, adopting the DeepLabV3+ architecture with a ViT-B backbone. As shown in Table \ref{tab:s4}, FixMatch demonstrates the highest efficiency, requiring only 9.51 GB of GPU memory and 7.10 hours per epoch. In contrast, UniMatch and CorrMatch introduce significantly higher GPU memory consumption and training time. Moreover, the performance gap in fine-tuning results among models pre-trained with different S4 methods narrows significantly when large-scale data is used. Considering both pre-training cost and fine-tuning effectiveness, we adopt FixMatch as the default S4P method in our S5 framework, as it offers the best trade-off between performance and efficiency.

\subsection{Stage-wise Pre-training} 
S4P builds upon existing pretrained backbones and aims to further enhance the capability of RSFMs in downstream tasks. We also evaluate the performance of models trained from scratch using only S4P pre-training without MAE initialization. The detailed results are shown in Table~\ref{pretrn_ablation}. The experiments reveal that models lacking pretrained weight initialization perform poorly across all metrics. In contrast, introducing MAE \cite{MIM} pre-training leads to significant improvements in segmentation performance on the Vaihingen, LoveDA, and DIOR-R datasets. When the model is trained entirely from scratch and pre-trained solely with S4P, it shows clear improvements over random initialization. However, its overall performance still falls short of models initialized with MAE. This is likely due to the limitations imposed by a large number of noisy pseudo-labels, which hinder model convergence and representation learning. Notably, when S4P is applied on top of MAE initialization, the model achieves the best performance across all three datasets. These results validate that S4P serves as an effective complementary pre-training strategy, significantly enhancing the representation capability of RSFMs and leading to better performance in downstream applications.

\begin{table*}[h]
\centering
\caption{Comparison of fine-tuning performance on Vaihingen, LoveDA, and DIOR-R using different pre-training initializations.}
\small
\fontsize{8}{10.5}\selectfont
\setlength{\tabcolsep}{8.1pt}
\resizebox{0.65\linewidth}{!}{
\begin{tabular}{c|cc|ccc}
\hline
Model & MAE & S4P & Vaihingen & LoveDA & DIOR-R \\
\hline
\multirow{4}{*}{ViT-B + UperNet} 
& - & - & 65.72 & 35.41 & 49.78 \\
& \checkmark & - & 78.27 & 52.47 & 68.02 \\
& - & \checkmark & 74.03 & 44.28 & 52.99 \\
& \checkmark & \checkmark & \textbf{79.93} & \textbf{54.51} & \textbf{69.56} \\
\hline
\end{tabular}
}
\label{pretrn_ablation}
\end{table*}

\begin{table*}[h]
\centering
\caption{Comparison of fine-tuning results after pre-training on different labeled datasets using the UperNet framework with ViT-B backbone.}
\small
\fontsize{8}{10.5}\selectfont
\setlength{\tabcolsep}{8.1pt}
\resizebox{0.9\linewidth}{!}{
\begin{tabular}{clcccccc}
\hline
Labeled Dataset & Method & Backbone & Vaihingen & Potsdam & LoveDA & OpenEarthMap \\
\hline
- & MAE & ViT-B & 78.27 & 91.58 & 52.47 & 66.23 \\
iSAID & MAE + S4P & ViT-B & \textbf{79.93} & 92.24 & \textbf{54.51} & \textbf{67.35} \\
LoveDA & MAE + S4P & ViT-B & 79.18 & 92.01 & - & 66.13 \\
OpenEarthMap & MAE + S4P & ViT-B & 79.79 & \textbf{92.37} & 54.01 & - \\
\hline
\end{tabular}}
\label{tab:pretrain_datasets}
\end{table*}

\subsection{Pretraining with Other Labeled Datasets}
Based on the experimental results presented in Table~\ref{tab:pretrain_datasets}, pre-training with various labeled datasets followed by fine-tuning generally yields effective performance improvements. Nonetheless, the results are influenced by factors such as dataset size and annotation quality. Among the evaluated datasets, LoveDA shows the lowest performance, likely due to its smaller scale or inferior annotation quality; OpenEarthMap yields moderate results, while iSAID consistently outperforms the others across most benchmarks, demonstrating its superiority as a labeled pre-training source in our experimental setting.

\subsection{Evaluation on Natural Image Segmentation Benchmarks}
To further validate the effectiveness of the proposed S4P on natural images, we conduct experiments following the UniMatch V2 \cite{UnimatchV2} setting. We use ADE20K \cite{ade20k} as the labeled dataset, which contains 150 semantic categories and covers diverse indoor and outdoor scenes. In addition, we utilize unlabeled data from COCO \cite{COCO}, Cityscapes \cite{cityscapes}, and Pascal VOC \cite{pascal} (excluding their validation and test splits) for pre-training. Table~\ref{tab:natural_seg} summarizes the segmentation results on Cityscapes \cite{cityscapes}, COCO \cite{COCO}, and Pascal VOC \cite{pascal} using different pre-training strategies. Compared to MAE*, S4P yields consistent improvements across all datasets, achieving gains of +1.20\%, +2.08\%, and +2.86\% on Cityscapes, COCO, and Pascal VOC, respectively, when using the ViT-B backbone with UperNet. These results indicate that S5 not only enhances the representation capability of models in RSFMs, but also improves generalization in the natural image domain, further demonstrating its strong transferability across diverse vision tasks.

\subsection{Visualization of S5 Results on Downstream Tasks}
\begin{figure*}[t] 
\centering
\includegraphics[width=0.9\textwidth]{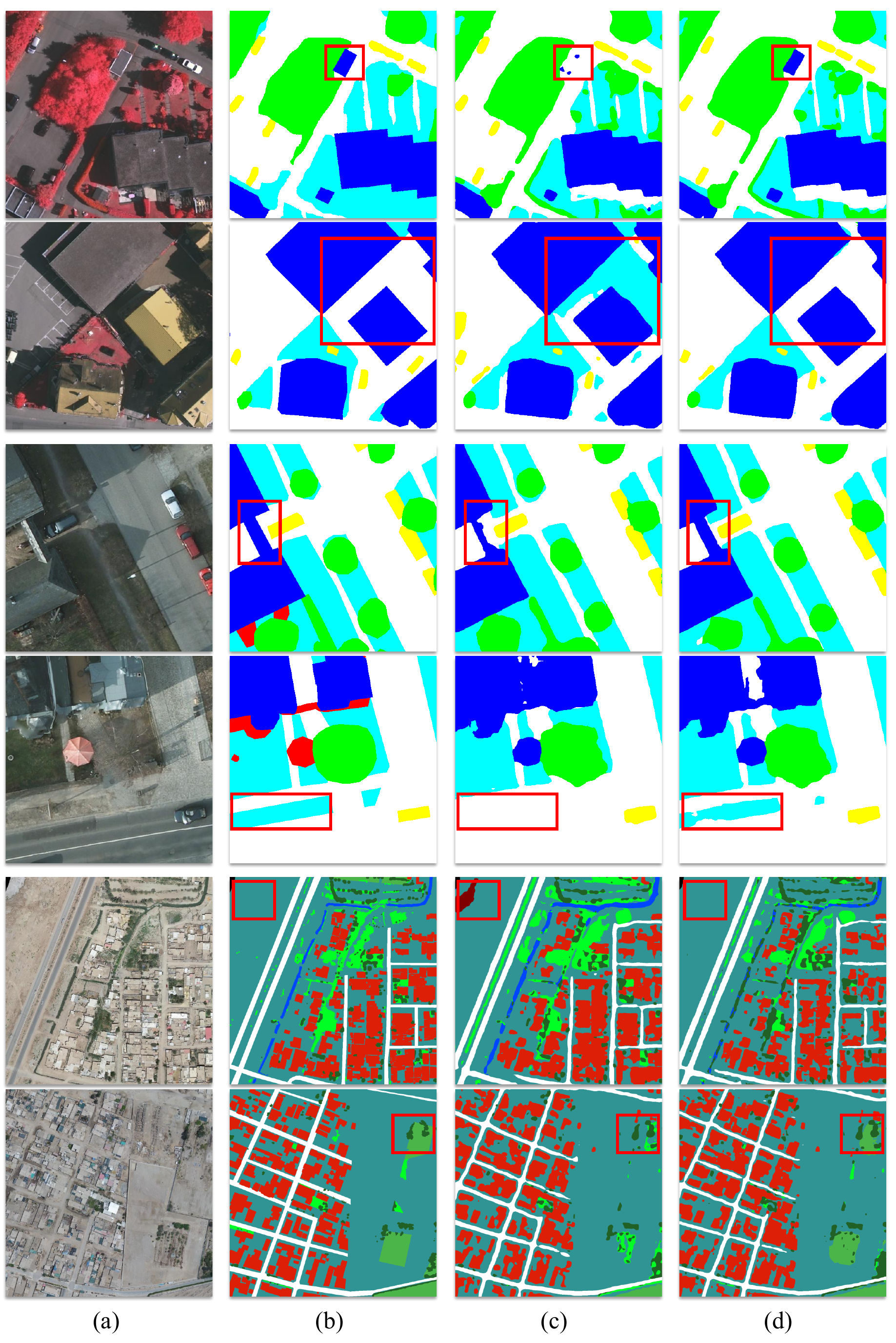}
\caption{Visualization of S5 predictions. Every two rows from top to bottom correspond to the Vaihingen, Potsdam, and OpenEarthMap datasets, respectively. (a) Image. (b) Ground truth. (c) Predictions from MAE. (d) Predictions from S5.
}
\label{fig:seg}
\end{figure*}

\begin{figure*}[t] 
\centering
\includegraphics[width=0.9\textwidth]{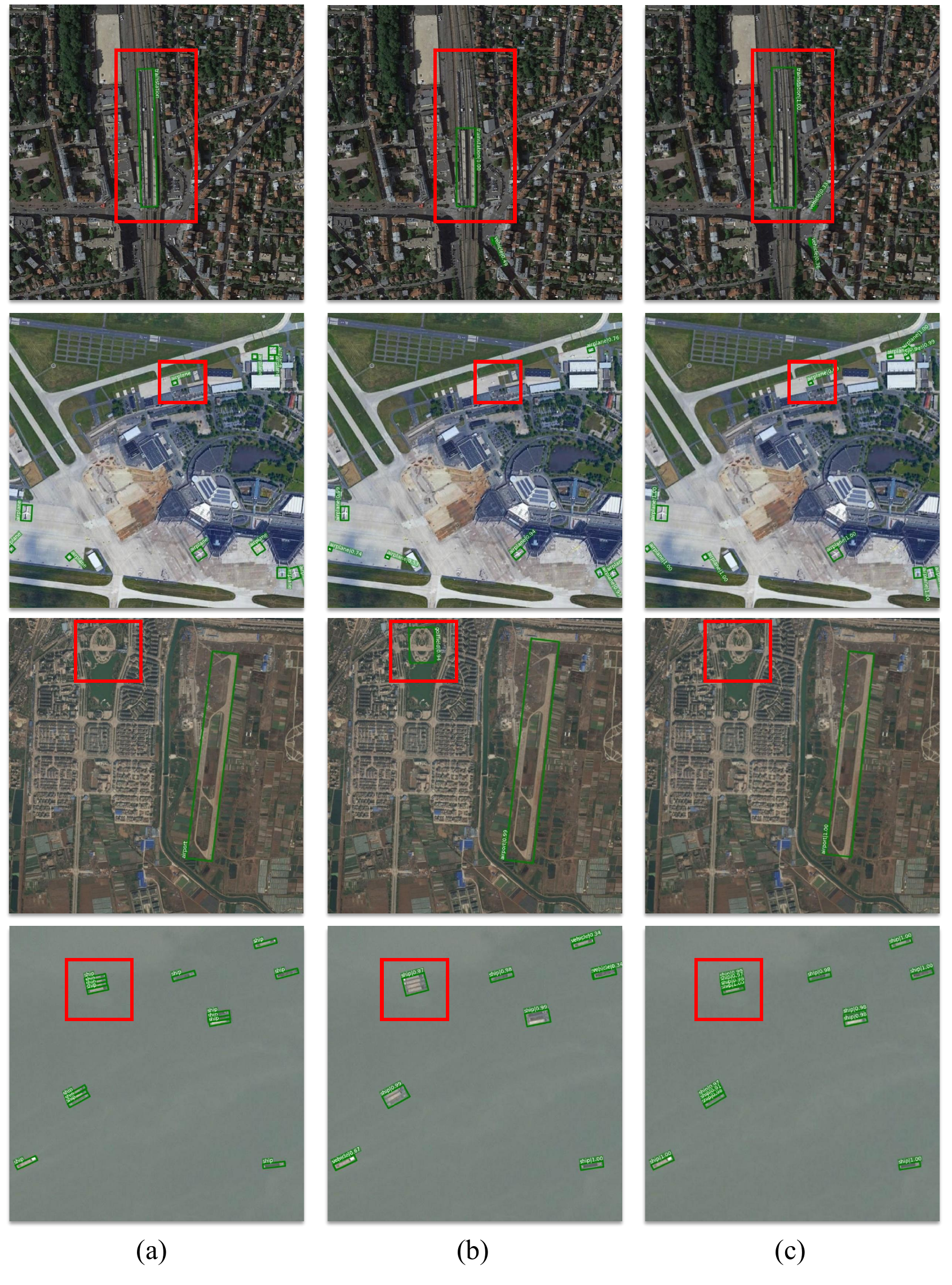}
\caption{
Visualization of S5 predictions on DIOR-R dataset. (a) Ground truth.(b) Predictions from MAE. (c) Predictions from S5.
}
\label{fig:det}
\end{figure*}

Considering that the LoveDA and DOTA-v2.0 test sets do not provide ground truth annotations, we present visual comparisons on the semantic segmentation benchmarks Vaihingen, Potsdam, and OpenEarthMap, as well as the object detection dataset DIOR-R. We use ViT-H as the backbone for all predictions. The Figure \ref{fig:seg} and \ref{fig:det} show that, compared to MAE, S5 produces predictions that are more consistent with the ground truth in both segmentation and detection tasks, ensuring higher accuracy. In object detection, our method demonstrates strong performance in identifying small and overlapping targets; in segmentation, it effectively captures and delineates the primary land cover types across large areas in RS images.

\begin{table*}[t]
\centering
\caption{Segmentation performance (mIoU) on natural image benchmarks using different pre-training methods. Results are reported on Cityscapes \cite{cityscapes}, COCO \cite{COCO}, and Pascal VOC \cite{pascal} using ViT-B \cite{ViT} as the backbone with UperNet \cite{UperNet}.  Where MAE* refers to the weights pre-trained on ImageNet \cite{ImageNet}.}
\label{tab:natural_seg}
\small
\fontsize{8}{10.5}\selectfont
\setlength{\tabcolsep}{8.1pt}
\resizebox{0.8\linewidth}{!}{
\begin{tabular}{lllccc}
\hline
Method & Pretrain & Backbone & Cityscapes & COCO & Pascal VOC \\
\hline
UperNet & MAE* & ViT-B  & 77.81 & 54.13 & 77.97 \\
UperNet & MAE* + S4P             & ViT-B & \textbf{79.01} & \textbf{56.21} & \textbf{80.83} \\
\hline
\end{tabular}}
\end{table*}

\section{Datasheet}

\subsection{Motivation}

\noindent \textbf{1. For what purpose was the dataset created? Was there a specific task in mind? Was there a specific gap that needed to be filled? Please provide a description.}

\textbf{A1:} RS4P-1M is designed to advance research in the field of RSFMs. Due to the complexity of pixel annotation in RS images, the field still lacks large-scale RS segmentation datasets, hindering the implementation of RS segmentation pre-training. As a large-scale RS segmentation pre-training dataset with a capacity reaching the million level, RS4P-1M, when combined with S4 method, can effectively bridge this gap.

\noindent \textbf{2. Who created this dataset (e.g., which team, research group) and on behalf of which entity (e.g., company, institution, organization)?}

\textbf{A2:} RS4P-1M is created by the authors.

\noindent \textbf{3. Who funded the creation of the dataset? If there is an associated grant, please provide the name of the grantor and the grant name and number.}

\textbf{A3:} N/A.

\subsection{Composition}

\noindent \textbf{1. What do the instances that comprise the dataset represent (e.g., documents, photos, people, countries)? Are there multiple types of instances(e.g., movies, users, and ratings; people and interactions between them; nodes and edges)? Please provide a description.}

\textbf{A1:} RS4P-1M consists of 16 categories, with each sample comprising a remote sensing image and its corresponding pixel-level semantic labels.

\noindent \textbf{2. How many instances are there in total (of each type, if appropriate)?}

\textbf{A2:} RS4P-1M has 1,000,000 images.

\noindent \textbf{3. Does the dataset contain all possible instances or is it a sample (not necessarily random) of instances from a larger set? If the dataset is a sample, then what is the larger set? Is the sample representative of the larger set (e.g., geographic coverage)? If so, please describe how this representativeness was validated/verified. If it is not representative of the larger set, please describe why not (e.g., to cover a more diverse range of instances, because instances were withheld or unavailable).}

\textbf{A3:} RS4P-1M is a real-world sample dataset of global ground objects, containing pixel-level classification annotations. It is the largest dataset in the field of high-resolution remote sensing segmentation, with a scale reaching the million level compared to other high-resolution remote sensing segmentation pre-training datasets.

\noindent \textbf{4. What data does each instance consist of? “Raw” data (e.g., unprocessed text or images)or features? In either case, please provide a description.}

\textbf{A4:} Each instance consists of one land object with its pixel-level semantic annotations and the unprocessed image data.

\noindent \textbf{5. Is there a label or target associated with each instance? If so, please provide a description.}

\textbf{A5:} Yes. Each target is associated with pixel-level semantic labels lying in the corresponding *.png image.

\noindent \textbf{6. Is any information missing from individual instances? If so, please provide a description, explaining why this information is missing (e.g., because it was unavailable). This does not include intentionally removed information, but might include, e.g., redacted text.}

\textbf{A6:} Yes. A limited number of instances may exhibit incomplete masks, as the labels are obtained by S4P pre-trained UperNet with ViT-H backbone.

\noindent \textbf{7. Are relationships between individual instances made explicit (e.g., users’ movie ratings, social network links)? If so, please describe how these relationships are made explicit.}

\textbf{A7:} Yes. The instances' information is stored in the *.png images, and different instances can be clearly by pixel positions and filenames.

\noindent \textbf{8. Are there recommended data splits (e.g., training, development/validation, testing)? If so, please provide a description of these splits, explaining the rationale behind them.}

\textbf{A8:} Yes. RS4P-1M is designed to be used in conjunction with the iSAID dataset. Specifically, the iSAID training set (38,936 annotated images) together with RS4P-1M serve as the training data for S4P, while the officially defined iSAID validation set (21,183 annotated images) is used to evaluate the correctness of S4P.

\noindent \textbf{9. Are there any errors, sources of noise, or redundancies in the dataset? If so, please provide a description.}

\textbf{A9:} Since most of the images in RS4P-1M are raw and unlabeled, we used the S4 pre-trained UperNet to generate pixel-level pseudo-labels for them, which may contain some noisy annotations. 

\noindent \textbf{10. Is the dataset self-contained, or does it link to or otherwise rely on external resources (e.g., websites, tweets, other datasets)? If it links to or relies on external resources, a) are there guarantees that they will exist, and remain constant, over time; b) are there official archival versions of the complete dataset (i.e., including the external resources as they existed at the time the dataset was created); c) are there any restrictions (e.g., licenses, fees) associated with any of the external resources that might apply to a future user? Please provide descriptions of all external resources and any restrictions associated with them, as well as links or other access points, as appropriate.}

\textbf{A10:} The RS4P-1M dataset is composed of several publicly available datasets, including MillionAID~\cite{MillionAID}, SAMRS~\cite{SAMRS} and STAR~\cite{STAR}. These datasets are publicly accessible and can be downloaded from their respective websites. We sincerely appreciate the significant contributions of their authors to the research community.

\noindent \textbf{11. Does the dataset contain data that might be considered confidential (e.g., data that is protected by legal privilege or by doctorpatient confidentiality, data that includes the content of individuals non-public communications)? If so, please provide a description.}

\textbf{A11:} No.

\noindent \textbf{12. Does the dataset contain data that, if viewed directly, might be offensive, insulting, threatening, or might otherwise cause anxiety? If so, please describe why.}

\textbf{A12:} No.

\subsection{Collection Process}

\noindent \textbf{1. How was the data associated with each instance acquired? Was the data directly observable (e.g., raw text, movie ratings), reported by subjects (e.g., survey responses), or indirectly inferred/derived from other data (e.g., part-of-speech tags, model-based guesses for age or language)? If data was reported by subjects or indirectly inferred/derived from other data, was the data validated/verified? If so, please describe how.}

\textbf{A1:} The data associated with each instance are directly observable, as they are stored in the common png format and can be easily viewed via \url{https://pypi.org/project/Pillow/}{Python Imaging Library} or \url{https://opencv.org/}{Open Source Computer Vision Library}.

\noindent \textbf{2. What mechanisms or procedures were used to collect the data (e.g., hardware apparatus or sensor, manual human curation, software program, software API)? How were these mechanisms or procedures validated?}

\textbf{A2:} The images in RS4P-1M come from dataset publicly available datasets described above, which can be directly downloaded from their websites.

\noindent \textbf{3. If the dataset is a sample from a larger set, what was the sampling strategy (e.g., deterministic, probabilistic with specific sampling probabilities)?}

\textbf{A3:} No.

\noindent \textbf{4. Who was involved in the data collection process (e.g., students, crowdworkers, contractors) and how were they compensated (e.g., how much were crowdworkers paid)?}

\textbf{A4:} The authors.

\noindent \textbf{5. Over what timeframe was the data collected? Does this timeframe match the creation timeframe of the data associated with the instances (e.g., recent crawl of old news articles)? If not, please describe the timeframe in which the data associated with the instances was created.}

\textbf{A5}: Data collection took approximately 3 days, and annotation took around 5 days. Independent processing via programming was required, including clipping, standardizing filenames and formats, and converting label formats. Finally, we used the S4 pre-trained UperNet \cite{UperNet} with the ViT-H \cite{ViT} backbone to generate labels.

\subsection{Preprocessing/cleaning/labeling}

\noindent \textbf{1. Was any preprocessing/cleaning/labeling of the data done (e.g., discretization or bucketing, tokenization, part-of-speech tagging, SIFT feature extraction, removal of instances, processing of missing values)? If so, please provide a description. If not, you may skip the remainder of the questions in this section.}

\textbf{A1:} We crop all images to a size of 512 × 512 and apply  entropy-based filtering and semantic diversity expansion strategy to select images from the SAMRS, MillionAID, and STAR datasets.

\noindent \textbf{2. Was the “raw” data saved in addition to the preprocessed/cleaned/labeled data (e.g., to support unanticipated future uses)? If so, please provide a link or other access point to the “raw” data.}

\textbf{A2:} No.

\noindent \textbf{3. Is the software used to preprocess/clean/label the instances available? If so, please provide a link or other access point.}

\textbf{A3:} We use the \url{https://pypi.org/project/Pillow/}{Python Imaging Library} for cropping.

\subsection{Uses}

\noindent \textbf{1. Has the dataset been used for any tasks already? If so, please provide a description.}

\textbf{A1:} No.

\noindent \textbf{2. Is there a repository that links to any or all papers or systems that use the dataset? If so, please provide a link or other access point.}

\textbf{A2:} A link will be provided upon acceptance of this paper.

\noindent \textbf{3. What (other) tasks could the dataset be used for?}

\textbf{A3:} RS4P-1M can be used for the research of supervised or self-supervised pre-training of RS semantic segmentation models. 

\noindent \textbf{4. Is there anything about the composition of the dataset or the way it was collected and preprocessed/cleaned/labeled that might impact future uses? For example, is there anything that a future user might need to know to avoid uses that could result in unfair treatment of individuals or groups (e.g., stereotyping, quality of service issues) or other undesirable harms (e.g., financial harms, legal risks) If so, please provide a description. Is there anything a future user could do to mitigate these undesirable harms?}

\textbf{A4:} No.

\noindent \textbf{5. Are there tasks for which the dataset should not be used? If so, please provide a description.}

\textbf{A5:} No.

\subsection{Distribution}

\noindent \textbf{1. Will the dataset be distributed to third parties outside of the entity (e.g., company, institution, organization) on behalf of which the dataset was created? If so, please provide a description.}

\textbf{A1:} Yes. The dataset will be publicly available.

\noindent \textbf{2. How will the dataset will be distributed (e.g., tarball on website, API, GitHub)? Does the dataset have a digital object identifier (DOI)?}

\textbf{A2:} It will be publicly available on the project website.

\noindent \textbf{3. When will the dataset be distributed?}

\textbf{A3:} The dataset will be distributed once the paper is accepted after peer review.

\noindent \textbf{4. Will the dataset be distributed under a copyright or other intellectual property (IP) license, and/or under applicable terms of use (ToU)? If so, please describe this license and/or ToU, and provide a link or other access point to, or otherwise reproduce, any relevant licensing terms or ToU, as well as any fees associated with these restrictions.}

\textbf{A4:} It will be distributed under the MIT license.

\noindent \textbf{5. Have any third parties imposed IP-based or other restrictions on the data associated with the instances? If so, please describe these restrictions, and provide a link or other access point to, or otherwise reproduce, any relevant licensing terms, as well as any fees associated with these restrictions.}

\textbf{A5:} No.

\noindent \textbf{6. Do any export controls or other regulatory restrictions apply to the dataset or to individual instances? If so, please describe these restrictions, and provide a link or other access point to, or otherwise reproduce, any supporting documentation.}

\textbf{A6:} No.

\subsection{Maintenance}

\noindent \textbf{1. Who will be supporting/hosting/maintaining the dataset?}

\textbf{A1:} The authors.

\noindent \textbf{2. How can the owner/curator/manager of the dataset be contacted (e.g., email address)?}

\textbf{A2:} They can be contacted via email available on the project website.

\noindent \textbf{3. Is there an erratum? If so, please provide a link or other access point.}

\textbf{A3:} No. 

\noindent \textbf{4. Will the dataset be updated (e.g., to correct labeling errors, add new instances, delete instances)? If so, please describe how often, by whom, and how updates will be communicated to users (e.g., mailing list, GitHub)?}

\textbf{A4:} No.

\noindent \textbf{5. Will older versions of the dataset continue to be supported/hosted/maintained? If so, please describe how. If not, please describe how its obsolescence will be communicated to users.}

\textbf{A5:} N/A.

\noindent \textbf{6. If others want to extend/augment/build on/contribute to the dataset, is there a mechanism for them to do so? If so, please provide a description. Will these contributions be validated/verified? If so, please describe how. If not, why not? Is there a process for communicating/distributing these contributions to other users? If so, please provide a description.}

\textbf{A6:} N/A.

\end{document}